\documentclass{article}

\usepackage[numbers]{natbib}

\usepackage[preprint]{neurips_2026}

\usepackage[utf8]{inputenc} 
\usepackage[T1]{fontenc}    
\usepackage{hyperref}       
\usepackage{url}            
\usepackage{booktabs}       
\usepackage{amsfonts}       
\usepackage{nicefrac}       
\usepackage{microtype}      
\usepackage{xcolor}         

\usepackage{graphicx}
\usepackage{subcaption}

\usepackage[most]{tcolorbox}
\usepackage{amsmath}
\usepackage{amssymb}
\usepackage{mathtools}
\usepackage{amsthm}
\usepackage{algorithm}
\usepackage{algorithmic}

\DeclareMathOperator*{\argmax}{arg\,max}
\usepackage[capitalize,noabbrev]{cleveref}

\theoremstyle{plain}
\newtheorem{theorem}{Theorem}[section]
\newtheorem{proposition}[theorem]{Proposition}

\newtheorem{result}[theorem]{Result}
\theoremstyle{definition}

\theoremstyle{remark}

\usepackage[textsize=tiny]{todonotes}
\usepackage{bm}

\title{Unveiling Memorization--Generalization Coexistence:\\ 
  A Case Study on Arithmetic Tasks with Label Noise}

\author{
  Linyu Liu$^{1}$ \qquad Pinyan Lu$^{1,2}$ \\
  $^{1}$Taylor Lab, Huawei Technologies Co., Ltd. \\
  $^{2}$Key Laboratory of Interdisciplinary Research of Computation and Economics, \\
  Shanghai University of Finance and Economics \\
  \texttt{liulinyu001@gmail.com} \qquad \texttt{lu.pinyan@mail.shufe.edu.cn}
}

\begin{document}

\maketitle

\begin{abstract}
  Highly over-parameterized models can simultaneously memorize noisy labels and generalize well, yet how these behaviors coexist remains poorly understood. In this work, we investigate the underlying mechanisms of this coexistence using modular arithmetic tasks under heavy label noise. Through extensive experiments on two-layer neural networks, we find that larger models tend to generalize better under appropriate optimization and model configurations, while noisy labels are memorized faster than clean data. Over-parameterized models internally form a generalization structure, but its expression in the output is suppressed by the need to fit noisy labels. Remarkably, even with 80\% label noise, near-perfect test accuracy can be achieved by extracting this internal structure using frequency-based methods. We further propose a task-agnostic method to partition networks into generalization and memorization components. Although this subnetwork improves generalization, it is limited compared with frequency-based extraction, indicating that the generalization structure is distributed across neurons and motivating the development of new tools to retrieve generalizable knowledge from over-parameterized networks.
\end{abstract}

\section{Introduction}
Large Language Models succeed largely because increasing model capacity consistently improves generalization \citep{brown2020language, kaplan2020scaling,henighan2020scaling,hoffmann2022training}. In structured tasks, larger models exhibit emergent abilities to capture underlying rules from data \citep{wei2022emergent}. However, real-world training data is often contaminated with noise, such as mislabeled samples or factual inconsistencies. While the scaling paradigm suggests that larger models perform better in clean environments \citep{liu2025superposition}, it remains unclear how they handle corrupted data. Specifically, we need to understand the internal competition between two distinct components: generalization, which is the discovery of structural rules that apply to unseen data, and memorization, which is the fitting of specific noisy labels that do not reflect the underlying rule.

Analyzing this interaction in realistic tasks is challenging because generalization is often difficult to diagnose mechanistically. Unseen examples may share local patterns, semantic features, or distributional correlations with the training data, so high test accuracy does not by itself reveal whether the model has learned a true rule or exploited shortcuts. To address this, we use modular arithmetic tasks (such as addition and multiplication modulo $P$) as a controlled experimental environment. These tasks are ideal for several reasons. First, they provide a clear distinction between rule-based generalization and point-wise memorization \citep{pearce2023grokking}. Second, they exhibit grokking \citep{power2022grokking,nandaprogress,zhong2023clock}, a phenomenon where generalization happens long after the model fully memorizes the training set. Third, these tasks allow for rigorous analysis because they have known analytical solutions under certain setups \citep{morwanifeature, tian2025provable}. 

In this work, by introducing label noise into this setup, we can precisely track how a model resolves the conflict between capturing a mathematical rule and memorizing random noise. We conduct extensive experiments across diverse settings, covering various activation functions, optimizers, and regularization strengths. This broad approach allows us to determine if the benefits of scaling persist under extreme corruption and how the model internally organizes these two competing behaviors. We focus on three key questions:

\textsf{Q1.} Under what conditions does increasing model size improve generalization when labels are noisy?

\textsf{Q2.} Can models still recover internal structural rules under extreme label noise?

\textsf{Q3.} Is it possible to separate the neurons responsible for generalization from those used for memorization?

Through these experiments, we provide the following findings:
\begin{enumerate}
\item We show that scaling model size continues to improve generalization even with noisy data, provided that the optimization and regularization are properly configured.
\item We observe a counter-intuitive learning dynamics where noisy labels are often memorized faster than the clean ones at a very early stage.
\item We find that models can \textit{internally} recover the underlying algebraic rule under extreme label noise ($\sim 80\%$ noise), though these rules are suppressed by the need to interpolate noisy labels.
\item We demonstrate that partitioning the network has the potential to improve generalization, generalization and memorization remain deeply entangled at the neuronal level.
\end{enumerate}

These results highlight both the robustness and the limitation of over-parameterized learning under corruption: models may learn generalizable rules even when forced to memorize noise, but current neuron-level tools cannot fully extract this rule structure from memorization. The rest of the paper elaborates on these findings. Section~\ref{sec:experiment-setup} introduces the problem setup. Section~\ref{sec:double-descent} provides an overview of accuracy performance from both the end-of-training and dynamics perspectives (\textsf{Q1}). Section~\ref{sec:representation-of-generalization} mechanically analyzes the internal representation of the generalization (\textsf{Q2}). Section~\ref{sec:network-decomposition} examines whether memorization and generalization can be separated at the neuron level (\textsf{Q3}). We defer related work to Appendix~\ref{sec:related-work}.

\section{Problem Setup}\label{sec:experiment-setup}

We study a modular arithmetic task $a \circ b = c$ following \citet{power2022grokking}, where $a,b,c \in \{0,\dots,P-1\}$ and $\circ$ denotes a modular operation such as addition, subtraction, or multiplication. We focus primarily on modular addition and use other operations to test generality. The task is formulated as a $P$-class classification problem.

We train a two-layer neural network parameterized by $\theta$:
\begin{equation}
    \bm{f}_{\theta}(a,b)
    = \bm{W}\,\phi\!\left(\bm{U}\bm{e}_a + \bm{V}\bm{e}_b\right) + \bm{\mu},
    \label{equ:MLP-model}
\end{equation}
where $\theta=\{\bm{W},\bm{U},\bm{V},\bm{\mu}\}$. Here, $\bm{e}_a,\bm{e}_b\in\mathbb{R}^P$ are one-hot encodings of $a$ and $b$; $\bm{U},\bm{V}\in\mathbb{R}^{M\times P}$ are first-layer weights; $\bm{W}\in\mathbb{R}^{P\times M}$ is the second-layer weight matrix; $\bm{\mu}\in\mathbb{R}^{P}$ is the bias; and $M$ denotes the number of hidden neurons. The output $\bm{f}_{\theta}(a,b)\in\mathbb{R}^P$ is trained using cross-entropy loss to predict the one-hot label $\bm{e}_{a\circ b}$.

Unless stated otherwise, we fix $P=113$ following \citet{nandaprogress} and vary the model width as $M=2^k+1$ for $k\in\{4,\dots,12\}$, since ReLU models empirically generalize better at odd widths in these tasks \citep{pearce2023grokking}.

\paragraph{Dataset.}
The total dataset is
\[
\mathcal{D}_{\text{total}}
=\{(a,b,c): a,b\in\{0,\dots,P-1\},\ c=a\circ b\}.
\]
We randomly split $\mathcal{D}_{\text{total}}$ into training, validation, and test sets with ratios $50\%$, $20\%$, and $30\%$, respectively. In Appendix~\ref{sec:addition-robustly-double-descent}, we also vary the training ratio while fixing the test ratio.

To introduce label noise, we randomly select a subset
$\mathcal{D}_{\text{noise}}\subset\mathcal{D}_{\text{train}}$
with noise ratio $\alpha$, i.e., $\frac{|\mathcal{D}_{\text{noise}}|}{|\mathcal{D}_{\text{train}}|} = \alpha$. For each $(a,b,c)\in\mathcal{D}_{\text{noise}}$, we define the observed label
$\tilde{c}$ as
\[
\tilde{c} =
\text{Uniform}\big(\{0,\cdots,P-1\}\setminus\{c\}\big),
(a,b,c)\in \mathcal{D}_{\text{noise}}.
\]
The resulting noisy training subset is given by
\[
\tilde{\mathcal{D}}_{\text{noise}}
\coloneqq
\{(a,b,\tilde{c}) : (a,b,c)\in\mathcal{D}_{\text{noise}}\}.
\]

The noisy training set is
$\tilde{\mathcal{D}}_{\text{train}}
= \mathcal{D}_{\text{clean}} \cup \tilde{\mathcal{D}}_{\text{noise}}$.
All evaluations are performed on clean validation and test sets.

\paragraph{Evaluation.} To measure the performance of a model $\bm{f}$, we define its accuracy on a dataset $\mathcal{D}$ as
\[
\text{Acc}\left(\bm{f}, \mathcal{D}\right) = \frac{1}{\vert\mathcal{D}\vert} \sum_{(a,b,c)\in\mathcal{D}} \mathbb{I}\left(\hat{c}=c\right),
\]
where $\hat{c} = \argmax_{c'} \left[\bm{f}\left(a,b\right)\right]_{c'}$, and $\mathbb{I}\left(\cdot\right)$ is the indicator function.

\paragraph{Optimizers.}
We train models with cross-entropy loss using Adam, AdamW \citep{loshchilov2017decoupled}, and Muon \citep{jordan6muon}, with weight decay controlling regularization strength. We also experiment with SGD, but it is inefficient and requires careful hyperparameter tuning to achieve generalization (see Figure~\ref{fig:SGD_dynamics}).

Unless specified otherwise, all results are reported after 200{,}000 epochs of full-batch training. We apply a linear learning-rate warmup over the first 10 steps. The default setting uses ReLU activations with AdamW (learning rate $10^{-3}$, weight decay $0.1$, random seed 42) on the modular addition task.

\section{When Bigger Is Better}\label{sec:double-descent}
In this section, we investigate how the presence of label noise interacts with model capacity in modular arithmetic tasks. We approach this question from two complementary perspectives. First, we examine end-of-training performance, asking under which conditions—such as optimization and regularization choices—increasing model size consistently improves generalization despite noise. Second, we analyze the training dynamics to study whether the model learns clean labels and noisy labels at different rates.

\subsection{End-of-Training Performance}

\begin{figure}[!htbp]
  \centering
  \begin{subfigure}[t]{0.48\linewidth}
    \centering
    \includegraphics[width=\linewidth]{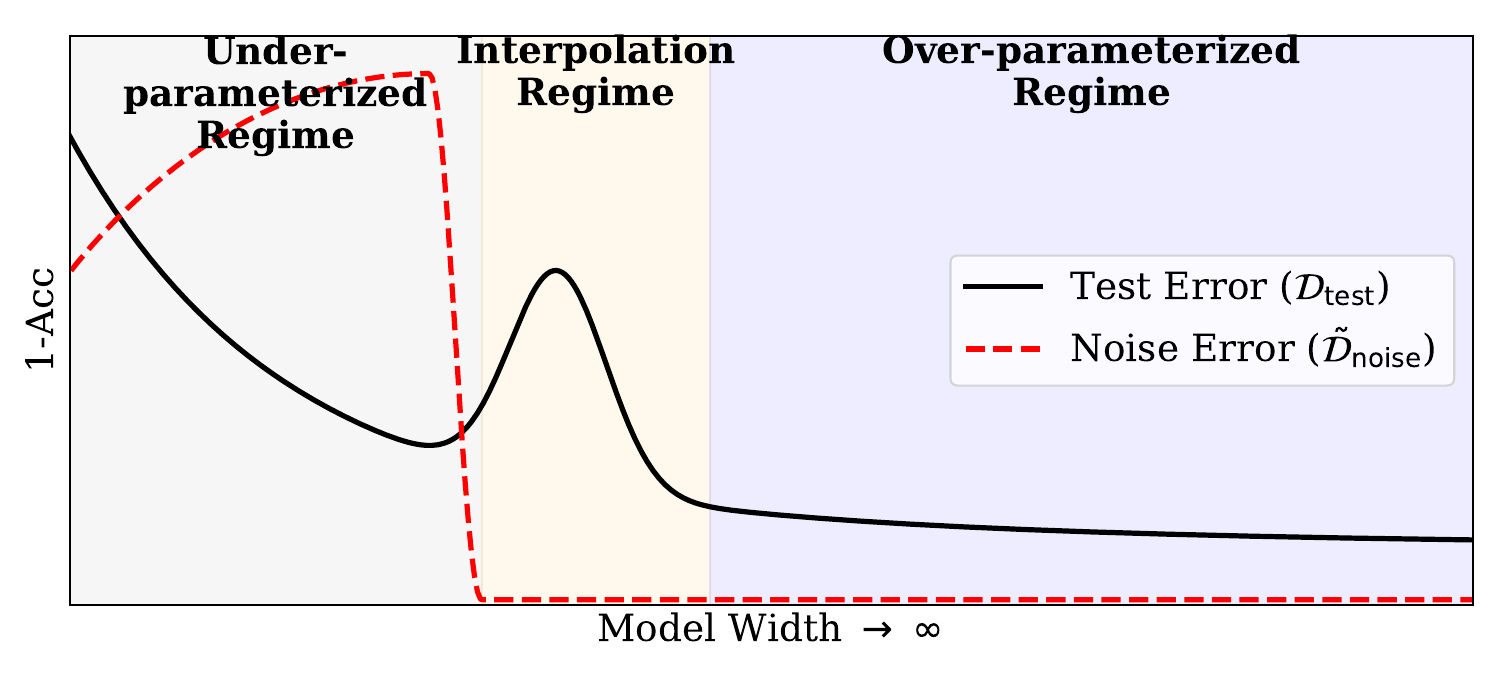}
    \caption{Illustration of double descent. Beyond the interpolation threshold, the over-parameterized regime corresponds to the second descent, where increasing model size leads to improved generalization, even after fitting all training samples, including label noise.}
    \label{fig:double_descent}
  \end{subfigure}
  \hfill
  \begin{subfigure}[t]{0.48\linewidth}
    \centering
    \includegraphics[width=\linewidth]{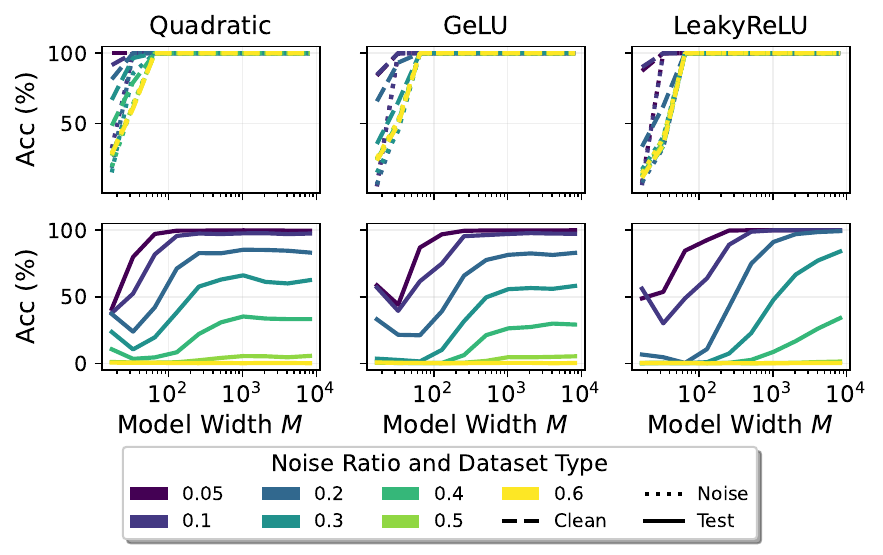}
    \caption{Performance across various activation functions. Generally, test accuracy is non-decreasing in the over-parameterized regime, but ceilings of test accuracy exist for the quadratic and GeLU activations.}
    \label{fig:across_activations}
  \end{subfigure}
  \caption{Model-wise double descent and performance across various activation functions.}
  \label{fig:double_descent_and_activations}
\end{figure}

\begin{figure*}[!htbp] 
  \centering
  \includegraphics[width=0.95\linewidth]{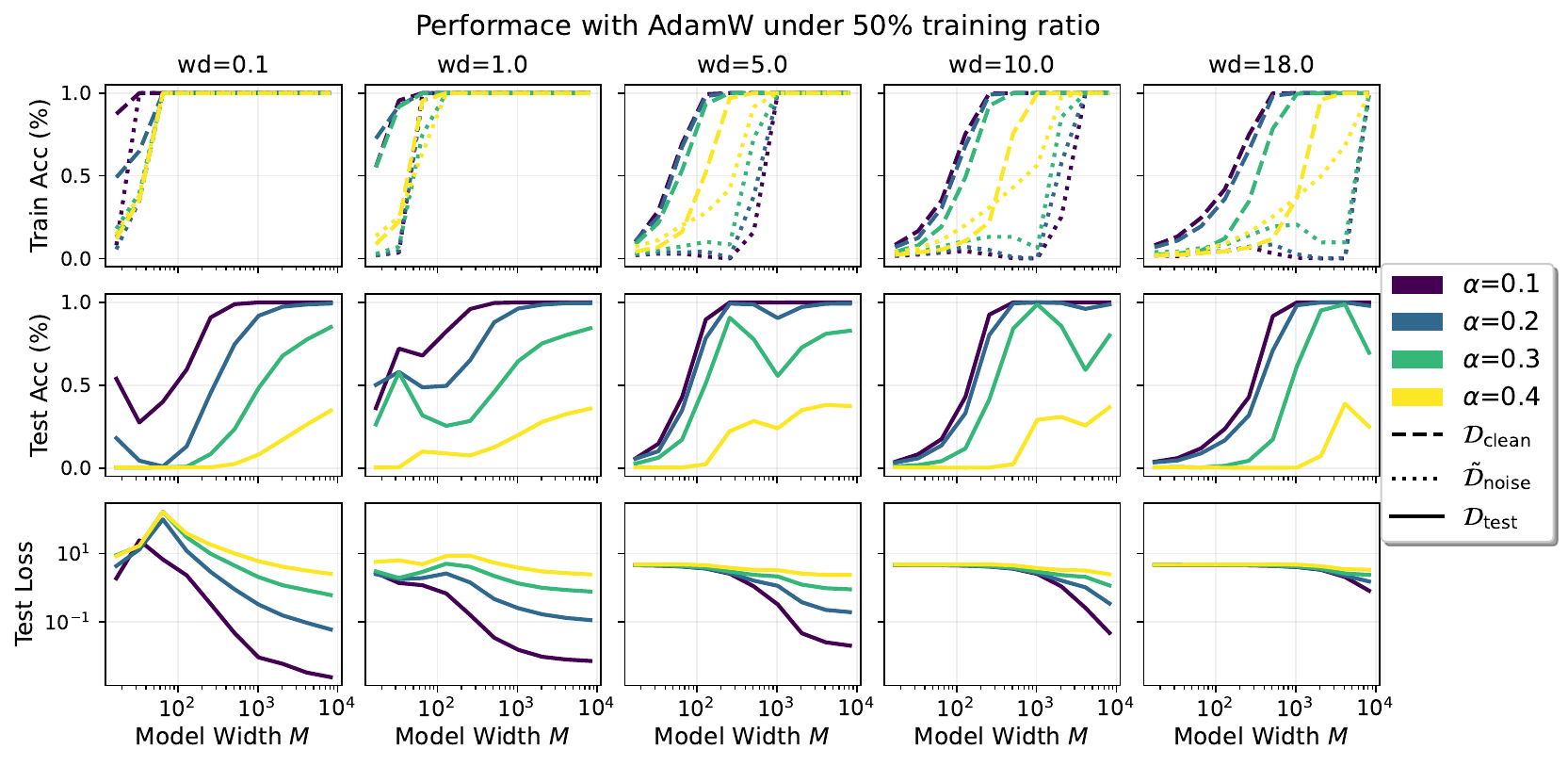}
  \caption{
Performance of ReLU models trained with AdamW under varying weight decay.
The first row shows training accuracy on clean ($\mathcal{D}_{\text{clean}}$) and noisy ($\tilde{\mathcal{D}}_{\text{noise}}$) data to characterize memorization.
The second row shows test accuracy, and the third row shows test cross-entropy loss.
Increasing weight decay shifts the interpolation threshold to larger sizes, delaying the second-descent regime.
}
  \label{fig:AdamW_wds}
\end{figure*}

We robustly observe the \textit{double descent} phenomenon of the test \textit{error} ($1-\text{test accuracy}$) across different activation functions, noise-to-training ratios, optimizers (including AdamW, Adam, and Muon), and arithmetic tasks (see the full results in Section \ref{sec:addition-robustly-double-descent}). 
As illustrated in Fig.~\ref{fig:double_descent}, \textit{double descent} refers to the non-monotonic test-error curve in which generalization first worsens near the interpolation threshold and then improves again as model size further increases.
In the over-parameterized regime, continually enlarging the model leads to non-decreasing test accuracy, while the threshold for entering this regime depends on the regularization strength. 
We summarize the observations in Result \ref{res:regularization}.

\begin{result}\label{res:regularization}
Double descent is consistently observed across varying setups:

1. Within the over-parameterized regime, bigger models generalize better.

2. Strengthening regularization will shift models from the over-parameterized regime toward the under-parameterized regime. (Figure~\ref{fig:AdamW_wds})

3. 3. Model sizes of test-accuracy saturation are activation-dependent (Figure~\ref{fig:across_activations})
\end{result}

Regularization does play a role in shaping the double-descent curve by controlling the effective model complexity (EMC) \citep{nakkiran2021deep}. Intuitively, stronger regularization restricts the set of functions that can be realized during training, effectively reduce the model's capacity. When the EMC is limited, learning signals for generalization and memorization compete more strongly. As a result, increasing regularization can be beneficial when memorization of label noise is undesirable.

Models with quadratic or GeLU activation functions exhibit clear ceilings in test accuracy as model size increases within the explored model size range (width $<10{,}000$, Figure \ref{fig:across_activations}). We include quadratic activations here because their analytical solutions are available (see Section~\ref{sec:representation-of-generalization}). One possible explanation for these ceilings is that the symmetry of these activations constrains how noise can be absorbed while preserving generalization. 
In contrast, ReLU activations store noise in a non-symmetric manner, while the representation of generalization remains symmetric (see Section~\ref{sec:representation-of-generalization}), allowing model capacity to be used more flexibly and leading to improved generalization.

We also observe that models are more confident (i.e., exhibit lower test cross-entropy loss) in the over-parameterized regime than in the under-parameterized regime. Larger models are generally more confident (Figure~\ref{fig:AdamW_wds}, bottom row). Cross-entropy loss reflects the confidence calibration of model predictions. In the under-parameterized regime, models often predict the true label even for corrupted samples. To align the predicted probability with the portion of correctness, the model assigns lower probabilities, resulting in higher uncertainty and higher test cross-entropy.
In contrast, in the over-parameterized regime, larger models can separate data with larger margins while still satisfying the same $\ell_2$ regularization constraint. Larger margins lead to more confident predictions and, consequently, lower test cross-entropy loss.

\textbf{Remarks on model misspecification.}
We note that some model architectures are unable to interpolate all noisy labels, in which case the over-parameterized regime does not strictly exist. For example, models with tied first-layer weights ($\bm{U}=\bm{V}$) cannot memorize asymmetric random noise where $a \circ b \neq b \circ a$. In such cases, test accuracy does not vary monotonically with model size, and selecting an appropriate regularization strength may be more effective than increasing model capacity. We provide a detailed discussion in Section~\ref{sec:model-misspecification}.

\subsection{Training Dynamics}
We observe a counter-intuitive phenomenon in the training dynamics: noisy labels are consistently memorized faster than clean labels at the initial training epochs, as summarized in Result~\ref{res:noise-memorize-faster}. While the learning rates for these two components are not drastically different, this inversion of the typical learning order is robust across various optimizers, including AdamW, Adam, and Muon (see Section~\ref{sec:additional-training-dynamics} for details). Even for SGD, which generally requires significantly longer timescales to achieve generalization, this trend remains consistent (Figure~\ref{fig:SGD_dynamics}).

\begin{result}\label{res:noise-memorize-faster}
In both under- and over-parameterized regimes, the model prioritizes the memorization of noisy labels over the fitting of clean labels.
\end{result}

\begin{figure}[!htbp]
  \centering
  \begin{subfigure}[t]{0.39\linewidth}
    \centering
    \includegraphics[width=\linewidth]{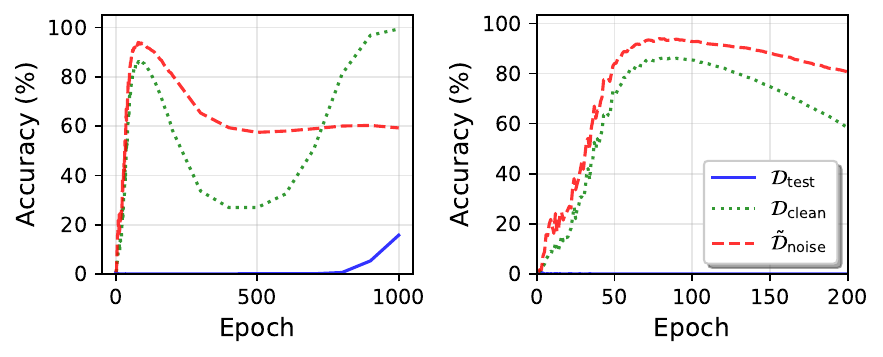}
    \caption{Under-parameterized regime.}
    \label{fig:training_dynamics_under}
  \end{subfigure}
  \hfill
  \raisebox{0.04\height}{\vrule width 0.4pt height 3.2cm}
  \hfill
  \begin{subfigure}[t]{0.58\linewidth}
    \centering
    \includegraphics[width=\linewidth]{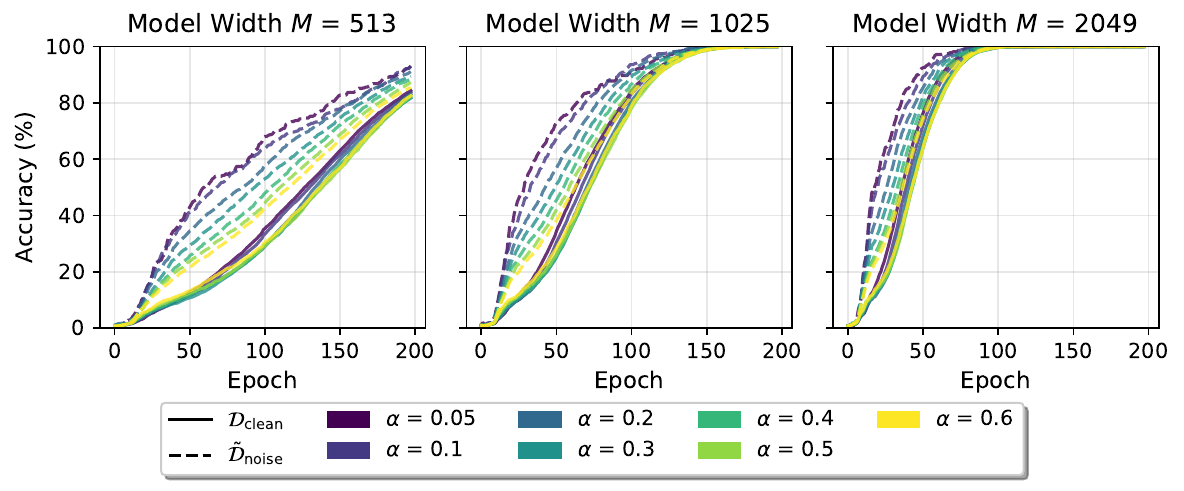}
    \caption{Over-parameterized regime.}
    \label{fig:training_dynamics_over}
  \end{subfigure}
  \caption{
  The faster memorization of noisy labels is observed in both the under- and
  over-parameterized regimes.
  \textbf{(a)} Under-parameterized learning dynamics with a zoom-in on the first
  200 epochs ($\alpha=0.3$, $M=2049$, Adam with weight decay $0.001$).
  \textbf{(b)} Over-parameterized learning dynamics under AdamW.
  In both cases, noisy-label accuracy rises earlier than clean-label accuracy.
  }
  \label{fig:training_dynamics}
\end{figure}

Selected learning curves are shown in Figure~\ref{fig:training_dynamics}.
The faster memorization of noisy labels appears in both regimes, but the
over-parameterized case in Figure~\ref{fig:training_dynamics_over} further
reveals two patterns: (i) increased model capacity accelerates the memorization
process, aligning with empirical findings in large language models
\citep{tirumala2022memorization}; and (ii) the gap between the memorization
speeds of noise and clean data becomes more pronounced at lower noise ratios
$\alpha$.

This observation stands in contrast to existing studies suggesting that models initially fit typical samples and ignore atypical ones, thereby making early stopping effective \citep{li2020gradient}. Our findings suggest a different early-stage dynamical behavior: under full-batch training, noisy labels can receive a larger per-sample optimization signal than clean labels, causing their accuracy to rise earlier. Importantly, this phenomenon is not restricted to two-layer networks on modular arithmetic tasks. In Section~\ref{sec:discussion-noise-memorize-faster}, we provide a gradient-based discussion and show that a similar early noisy-over-clean learning order also appears on MNIST with a three-layer neural network under full-batch training. Thus, while modular arithmetic provides a controlled setting where the behavior can be cleanly measured and analyzed, the phenomenon may reflect a broader property of early full-batch optimization under explicit label corruption. Additional results are provided in Section~\ref{sec:additional-training-dynamics}.

\section{Representation of Generalization}\label{sec:representation-of-generalization}

We address \textsf{Q2} by asking whether the apparent failure of ReLU networks under extreme label noise reflects the absence of a learned internal rule, or whether such a rule is learned but obscured by noisy outputs. We find evidence for the latter by separating the internal representation of the generalizable rule from that of memorization. Specifically, our main finding is that, although ReLU networks do not have the same expression as NN with quadratic activation, their generalization component closely matches the analytical solution known for quadratic activations under \textit{clean} data \citep{morwanifeature} (Equation \eqref{equ:exact-solution}). Thus, it suggests a way to separate the learned rule from memorized noise: retain the dominant frequency component aligned with this analytical structure and remove the residual components.

\begin{figure}[t]
    \centering
    \begin{subfigure}[t]{0.23\linewidth}
        \centering
        \includegraphics[width=\linewidth]{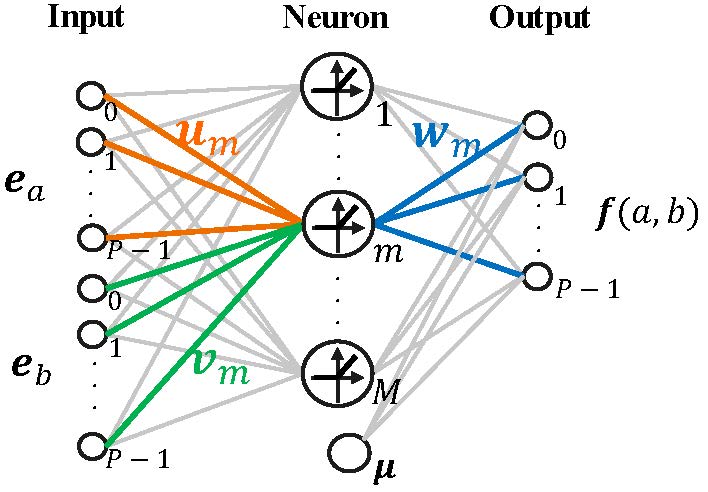}
        \caption{}
    \end{subfigure}
    \hspace{0.01\linewidth}
    \vrule width 0.8pt
    \hspace{0.01\linewidth}
    \begin{subfigure}[t]{0.28\linewidth}
        \centering
        \includegraphics[width=\linewidth]{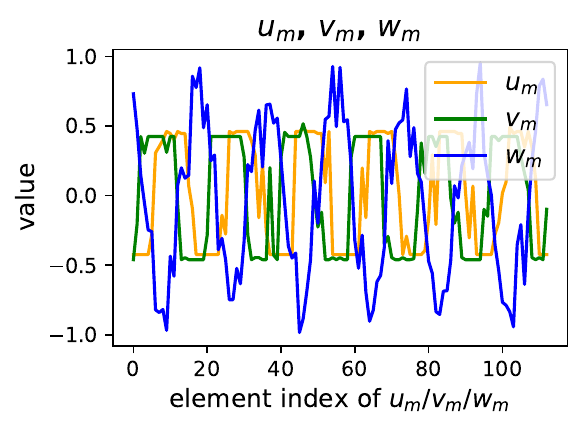}
        \caption{}
    \end{subfigure}
    \hspace{0.01\linewidth}
    \vrule width 0.8pt
    \hspace{0.01\linewidth}
    \begin{subfigure}[t]{0.4\linewidth}
        \centering
        \includegraphics[width=\linewidth]{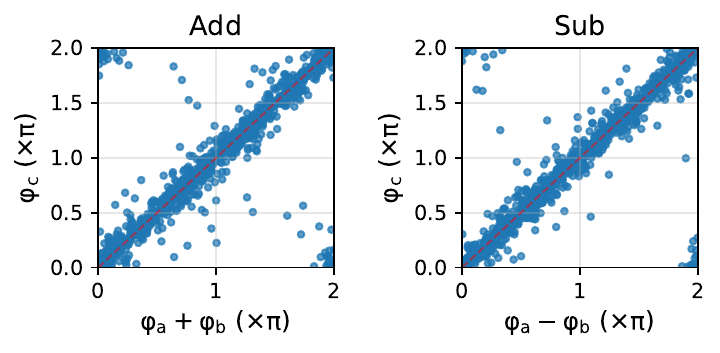}
        \caption{}
    \end{subfigure}
    \caption{\textbf{(a)} A neuron is represented by $(\bm{u}_m,\bm{v}_m,\bm{w}_m)$. \textbf{(b)} A representative neuron from a trained ReLU model exhibits periodicity. \textbf{(c)} Relationship of phases under $M=1025$ and $\alpha=0.3$. Each point represents a neuron, where $\varphi_a$, $\varphi_b$, and $\varphi_c$ are the phases of $\bm{u}^G$, $\bm{v}^G$, and $\bm{w}^G$, respectively.}
    \label{fig:neuron_phase_combined}
\end{figure}

\textbf{Analytical solution for NN with quadratic activation.} The two-layer NN \eqref{equ:MLP-model} can be equivalently written as the summation of the neurons:
\begin{equation*}
\bm{f}_{\theta} \left(a,b\right) = \sum_{m\in\mathcal{M}} \bm{w}_m \phi \left( \bm{u}^{\top}_m \bm{e}_a + \bm{v}^{\top}_m \bm{e}_b \right) + \bm{\mu},
\end{equation*}
where $\mathcal{M}=\left\{1, \cdots, M\right\}$. The triple, $\left(\bm{u}_m, \bm{v}_m, \bm{w}_m\right)$, represents the $m^{\text{th}}$ neuron. Here, $\bm{w}_m$ is the $m^{\text{th}}$ column vector of $\bm{W}$, and $\bm{u}^{\top}_m$ and $\bm{v}^{\top}_m$ are the $m^{\text{th}}$ row vectors of $\bm{U}$ and $\bm{V}$, respectively. Figure \ref{fig:neuron_phase_combined} (a) visualizes this decomposition. The margin-maximizing solution for modular addition with quadratic activation $\phi(z)=z^2$ takes the form~\citep{gromov2023grokking,morwanifeature}
\begin{equation}
u_{mi} = \lambda \cos(\tfrac{2\pi}{P}\omega_m i + \varphi^{(a)}_m),
  v_{mj} = \lambda \cos(\tfrac{2\pi}{P}\omega_m j + \varphi^{(b)}_m), 
  w_{mk} = \lambda \cos(\tfrac{2\pi}{P}\omega_m k + \varphi^{(c)}_m),
\label{equ:exact-solution}
\end{equation}
with $\varphi^{(a)}_m + \varphi^{(b)}_m = \varphi^{(c)}_m$ and $\omega_m \in \{1,\dots,\frac{P-1}{2}\}$ (Prop.~\ref{prop:analytical_solution}). This solution will achieve 100\% accuracy on all the clean samples. Empirically, ReLU models trained on clean data also develop a periodic pattern in $(\bm{u}_m,\bm{v}_m,\bm{w}_m)$ (Figure \ref{fig:neuron_phase_combined} (b)), but there is no explicit solution for them. 

\textbf{Frequency filtration separates rule learning from noise memorization.}
Drawing inspiration from the analytical solution of the NN with quadratic activation, we hypothesize that the generalization signal in a ReLU neuron is carried by its dominant frequency. We perform a Fourier-based decomposition on each hidden neuron $m \in \mathcal{M}$, characterized by the parameters $(\bm{u}_m, \bm{v}_m, \bm{w}_m)$. Specifically, we apply a filter to isolate the frequency $\omega$ with the maximum magnitude in the weight vector $\bm{w}_m$, and retrieve the same frequency component from $\bm{u}_m$ and $\bm{v}_m$. This dominant component is referred to as the generalization part, $(\bm{u}^G_m, \bm{v}^G_m, \bm{w}^G_m)$, while the remaining frequency components constitute the residual part, $(\bm{u}^R_m, \bm{v}^R_m, \bm{w}^R_m)$. The formal mathematical definitions of this filtration process are detailed in Appendix \ref{sec:frequency_filtration_details}. We subsequently construct two sub-networks: one composed of the dominant frequency components, denoted by $\left\{\bm{U}^G, \bm{V}^G, \bm{W}^G, \bm{\mu}\right\}$, and another consisting of the residuals, $\left\{\bm{U}^R, \bm{V}^R, \bm{W}^R, \bm{\mu}\right\}$. By evaluating these partitioned models separately, we quantify their respective contributions to memorization and generalization.

\begin{result}\label{res:ReLU-match-quadratic}
The generalization representation of ReLU models closely matches the analytical solution \eqref{equ:exact-solution} for modular addition and subtraction tasks.
\end{result}

\begin{figure}[htbp]
    \centering
    \begin{subfigure}[t]{0.47\linewidth}
        \centering
        \includegraphics[width=\textwidth]{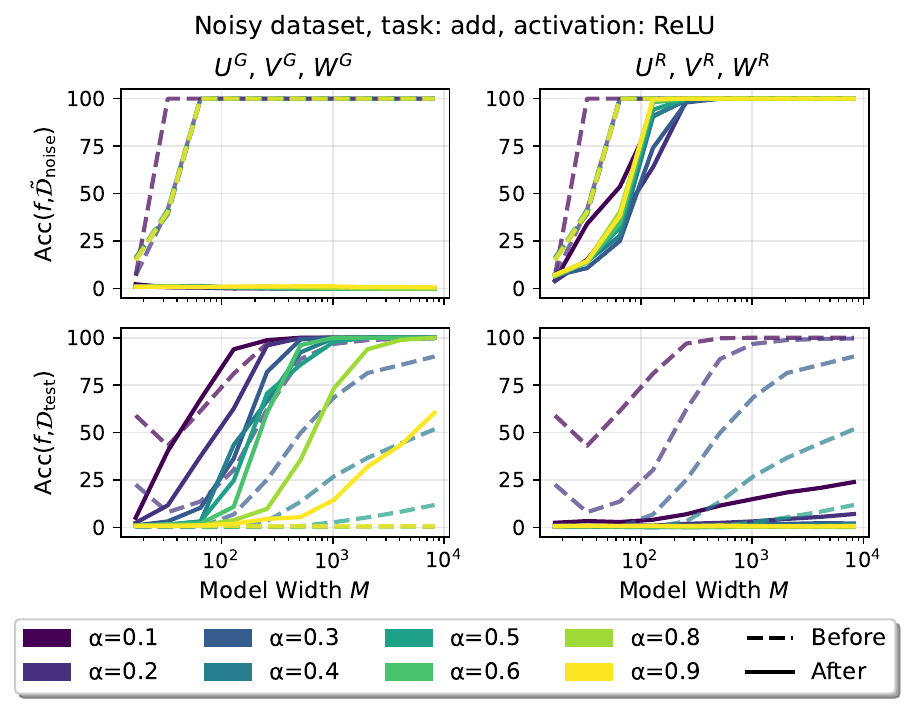}
    \end{subfigure}
    \hspace{0.015\linewidth}
    \vrule width 0.8pt
    \hspace{0.015\linewidth}
    \begin{subfigure}[t]{0.47\linewidth}
        \centering
        \includegraphics[width=\textwidth]{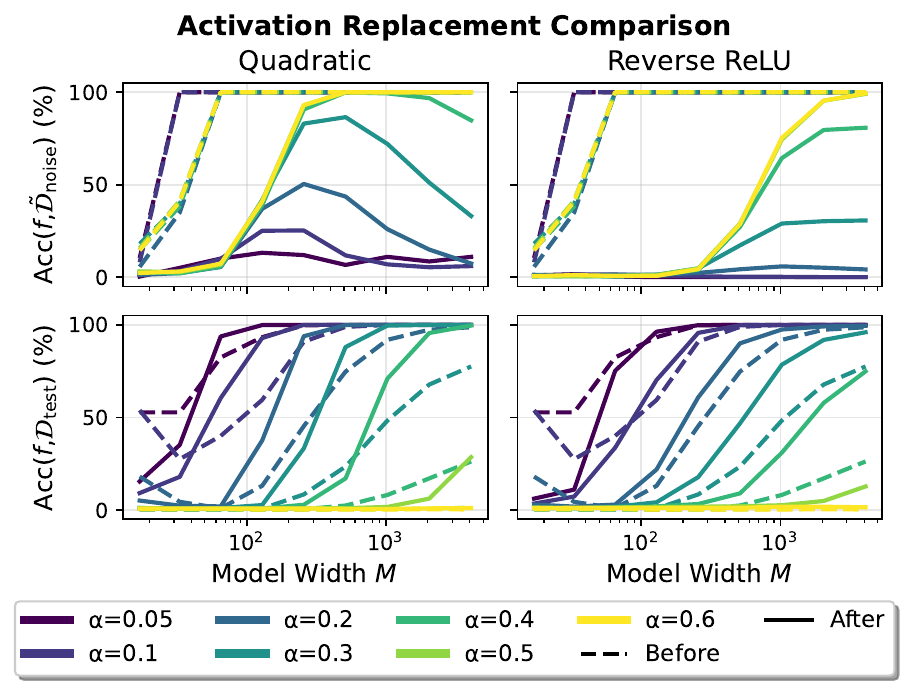}
    \end{subfigure}
    \caption{\textit{Left:} FF for ReLU networks. The dominant-frequency component $\{U^G,V^G,W^G\}$ recovers high test accuracy (after FF), while the residual component $\{U^R,V^R,W^R\}$ mainly retains noise memorization. \textit{Right:} Replacing the trained ReLU activation with quadratic or reverse ReLU preserves or even improves test accuracy and reduces noisy-label accuracy, suggesting that the learned weights already encode a rule representation compatible with other activation functions.}
    \label{fig:ff_activation_combined}
\end{figure}

Figure~\ref{fig:ff_activation_combined} (left) shows that the dominant-frequency sub-network achieves high test accuracy even under severe label noise, whereas the residual component is much more aligned with noisy-label memorization. This provides direct evidence that generalization and memorization coexist in the trained network but are encoded differently in frequency space.

The rule-separation effect of FF is surprisingly strong. Even when the model is trained with $80\%$ label noise, the dominant-frequency sub-network can recover near-perfect test accuracy, indicating that the model has internally learned the correct rule although its raw output is heavily corrupted by noise memorization. Conversely, the residual sub-network largely preserves the ability to fit noisy labels while contributing little to test generalization. This sharp contrast suggests that label noise does not simply prevent the model from learning the rule; instead, the rule is learned but masked by additional residual frequency components used for memorization. We further observe that FF also improves test accuracy for models trained on noise-free samples, for networks with quadratic activations, and for models on modular subtraction tasks, suggesting that dominant-frequency filtration is not merely removing noisy artifacts but extracting a general rule representation. The same frequency-based filtering idea can also be extended to a Transformer (see Appendix~\ref{sec:freq-filtration-additional}). 

\textbf{Additional evidence for the analytical structure.} We provide two complementary checks that the ReLU representation is coarsely aligned with the analytical solution. First, activation replacement shows that the learned rule does not rely on the exact shape of ReLU: after training, replacing ReLU with quadratic or reverse ReLU (defined as $\phi(x) = \max\{-x, 0\}$) preserves and often enhances generalization while reducing noisy-label accuracy (Figure~\ref{fig:ff_activation_combined} (right)). This is consistent with the hypothesis that the weights have already formed a symmetric rule representation. 

Second, the phases of the filtered dominant components match the phase relationship in the analytical solution. For addition, the phases satisfy $\varphi^{(a)}_m + \varphi^{(b)}_m \approx \varphi^{(c)}_m$; for subtraction, they satisfy $\varphi^{(a)}_m - \varphi^{(b)}_m \approx \varphi^{(c)}_m$ (Figure \ref{fig:neuron_phase_combined} (c)). These results, together with FF and activation replacement, support the same conclusion: ReLU networks generalize by forming a frequency-structured representation close to the quadratic analytical solution, enabling the disentanglement of the generalization and memorization in the internal representation. Additional experiment results are deferred to Appendix~\ref{sec:phase_additional}-\ref{sec:uniform_distribution_additional}.

The above evidences suggest that FF can disentangle memorization from generalization.
However, FF is inherently limited in two aspects:
(i) it does not yield a strict functional decomposition, since $\bm{f}_{\theta} \neq \bm{f}_{\{\bm{W}^G,\bm{U}^G,\bm{V}^G,\bm{\mu}\}} + \bm{f}_{\{\bm{W}^R,\bm{U}^R,\bm{V}^R,\bm{\mu}\}}$;
and (ii) it relies on prior knowledge of task-specific structure. This motivates the question of whether memorization and generalization can instead be separated in a task-agnostic manner (\textsf{Q3}). 

\section{Network Partitioning}\label{sec:network-decomposition}

We now address \textsf{Q3} by asking whether generalization and memorization can be separated at the neuron level. Specifically, we investigate whether a trained network can be partitioned into two neuron subsets: one subset $\mathcal{M}^G\subseteq \mathcal{M}$ responsible for generalization, and another subset $\mathcal{M}^R\subseteq \mathcal{M}$ responsible for memorizing label noise. Selecting $\mathcal{M}^G$ for generalization is equivalent to pruning the remaining neurons from the full network. For a selected neuron subset, the corresponding sub-networks are defined as
\begin{equation}
\small
\bm{f}_{\mathcal{M}^G}(a,b)
= \sum_{m\in\mathcal{M}^G} \bm{w}_m \phi\!\left(\bm{u}_m^{\top}\bm{e}_a + \bm{v}_m^{\top}\bm{e}_b\right) + \bm{\mu},\ 
\bm{f}_{\mathcal{M}^R}(a,b)
= \sum_{m\in\mathcal{M}^R} \bm{w}_m \phi\!\left(\bm{u}_m^{\top}\bm{e}_a + \bm{v}_m^{\top}\bm{e}_b\right) + \bm{\mu}.
\label{equ:fG-and-fR}
\end{equation}
We rank neurons using an importance score and select high-scoring neurons for generalization and low-scoring neurons for noise memorization. The full selection procedure is given in Algorithm \ref{alg:neuron_selection}.

\begin{algorithm}[htbp]
\caption{Neuron Selection for Network Decomposition}
\label{alg:neuron_selection}
\begin{algorithmic}[1]
\STATE {\textbf{Input:}} Network $\bm{f}_{\theta}$ with $M$ neurons $\mathcal{M}$; Metric $S(m)$; $\mathcal{D}_{\text{val}}$; $\tilde{\mathcal{D}}_{\text{noise}}$; Threshold $\tau$.
\STATE {\textbf{Initialize:}} $\mathcal{M}^G \leftarrow \emptyset, \mathcal{M}^R \leftarrow \emptyset$.
\STATE \COMMENT{// Step 1: Ranking neurons}
\STATE Calculate score $s_m = S(m)$ for each $m \in \mathcal{M}$.
\STATE Sort neurons descending: $\{m_{\pi(1)}, \dots, m_{\pi(M)}\}$ where $s_{m_{\pi(i)}} \ge s_{m_{\pi(j)}}$ for $i < j$.
\STATE \COMMENT{// Step 2: Selection for generalization}
\FOR{$k \in \{1, \dots, M\}$}
    \STATE $\mathcal{M}_k = \{m_{\pi(1)}, \dots, m_{\pi(k)}\}$;
    $A_k = \text{Acc}(\bm{f}_{\mathcal{M}_k}, \mathcal{D}_{\text{val}})$.
\ENDFOR
\STATE $k^G = \arg\max_{k} A_k$; 
$\gamma^G = k^G / M$ and $\mathcal{M}^G = \{m_{\pi(1)}, \dots, m_{\pi(k^G)}\}$.
\STATE \COMMENT{// Step 3: Minimal selection for noise memorization}
\FOR{$k \in \{1, \dots, M\}$}
    \STATE $\mathcal{M}'_k = \{m_{\pi(M-k+1)}, \dots, m_{\pi(M)}\}$. \COMMENT{Bottom-up}
    \IF{$\text{Acc}(\bm{f}_{\mathcal{M}'_k}, \tilde{\mathcal{D}}_{\text{noise}}) \ge \tau$}
        \STATE $k^R = k$, \textbf{break}.
    \ENDIF
\ENDFOR
\STATE $\gamma^R = k^R / M$ and $\mathcal{M}^R = \{m_{\pi(M-k^R+1)}, \dots, m_{\pi(M)}\}$.
\STATE {\textbf{Output:}} Ratios $\gamma^G, \gamma^R$; Selected neurons $\mathcal{M}^G$, $\mathcal{M}^R$; Sub-networks $\bm{f}^G=\bm{f}_{\mathcal{M}^G}$, $\bm{f}^R=\bm{f}_{\mathcal{M}^R}$.
\end{algorithmic}
\end{algorithm}

A natural importance score is the inverse participation ratio (IPR)~\citep{pastor2016distinct,girvin2019modern,gromov2023grokking,doshigrok}, proposed and applied in \citet{doshigrok}. It measures the periodicity of each neuron and is defined as
\begin{equation}
    \text{IPR}_m \coloneqq \left( \frac{\lVert \tilde{\bm{w}}_m \rVert_4}{\lVert \tilde{\bm{w}}_m \rVert_2} \right)^4, \label{equ:IPR}
\end{equation}
where $\lVert \cdot \rVert_p$ is the $\ell_p$-norm and $\tilde{\bm{w}}_m$ is the Fourier transform of $\bm{w}_m$. Equation~\eqref{equ:IPR} implies $\text{IPR}_m\in [1/M,1]$, where a larger value indicates stronger periodicity. This metric is well-suited for modular addition \citep{doshigrok}, whose generalizable solutions rely on periodic structures.

To avoid relying on a task-specific Fourier prior, we also propose a task-agnostic score, neuron strength (Str.), defined as
\begin{equation}
\text{Str}_{m} \coloneqq \lVert \bm{w}_m \rVert_{\infty} \phi\left(
\lVert \bm{u}_m \rVert_{\infty} + \lVert \bm{v}_m \rVert_{\infty}
\right),
\label{equ:margin-contribution-definition}
\end{equation}
where $\lVert \cdot \rVert_{\infty}$ denotes the $\ell_{\infty}$-norm. Intuitively, Str. measures the largest possible contribution of a neuron to the output margin. Unlike IPR, it does not assume periodicity and can therefore be applied to tasks such as modular multiplication, where periodicity in the original input domain is no longer a direct proxy for generalization.

\begin{result}
Neurons that contribute most to generalization typically exhibit higher neuron strength. Specifically, neurons with high IPR also attain high Str. in modular addition and subtraction. 
\label{res:IPR-STR-correlated}
\end{result}

\begin{figure}[t]
    \centering
    \begin{subfigure}[t]{0.5\linewidth}
        \centering
        \includegraphics[width=\linewidth]{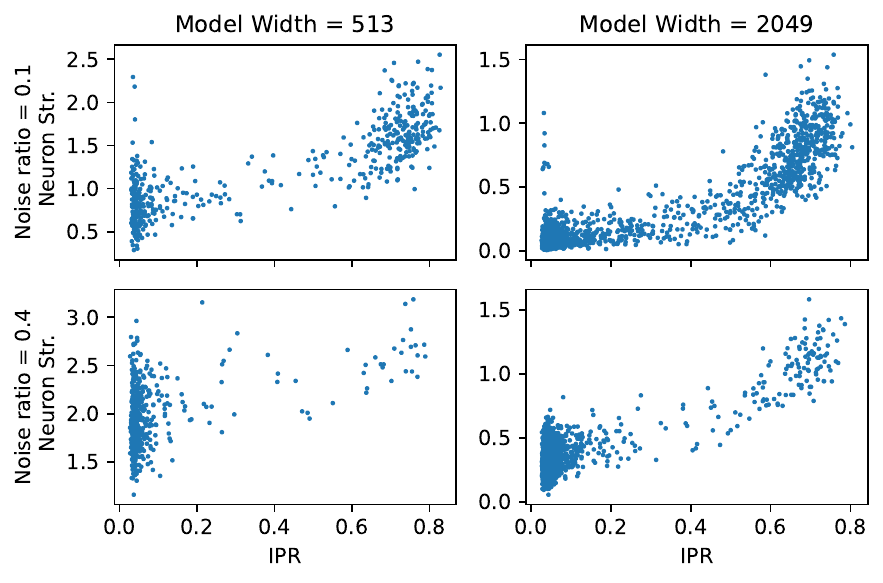}
    \end{subfigure}
    \hspace{0.02\linewidth}
    \begin{subfigure}[t]{0.45\linewidth}
        \centering
        \includegraphics[width=\linewidth]{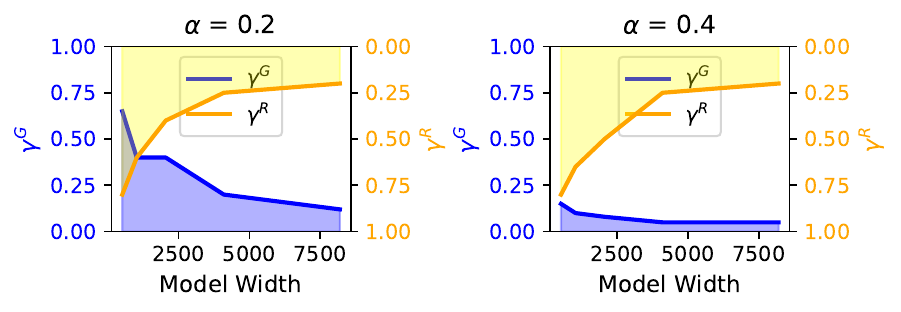}
    \end{subfigure}
    \caption{\textit{Left:} Neuron's IPR vs. Str. on the modular addition task. Neurons with stronger periodicity also tend to have larger Str., supporting Str. as a task-agnostic proxy for neuron importance. \textit{Right:} Neuron selection ratios using Str. on the modular addition task. Higher noise ratio leads to a smaller sub-network for generalization ($\gamma^G$). Two sub-networks have no overlapping neurons under sufficiently large model width.}
\label{fig:ipr_and_improvement_combined}
\end{figure}
\begin{result}
The sub-networks are not overlapped when model width is sufficiently large, i.e., $\gamma^G + \gamma^R \le 1$, where $\gamma^G = \frac{\vert\mathcal{M}^G\vert}{\vert\mathcal{M}\vert}$ and $\gamma^R = \frac{\vert\mathcal{M}^R\vert}{\vert\mathcal{M}\vert}$. (Figure \ref{fig:ipr_and_improvement_combined} right)
\end{result}

Figure~\ref{fig:ipr_and_improvement_combined} (left) shows that IPR and Str. are highly correlated. We also observe that a larger noise ratio reduces the number of high-IPR or high-Str. neurons. The optimal neuron ratio for generalization ($\gamma^G$) is also lower at a higher noise ratio (Figure~\ref{fig:ipr_and_improvement_combined} right). These may explain the deterioration of generalization under heavy label noise. Additional results in Appendix~\ref{sec:additional-network-decomposition} show that Str. yields performance comparable to IPR on addition and subtraction, while being more effective on modular multiplication, where IPR is less suitable.

\begin{figure}[htbp] 
  \centering
  \includegraphics[width=0.7\linewidth]{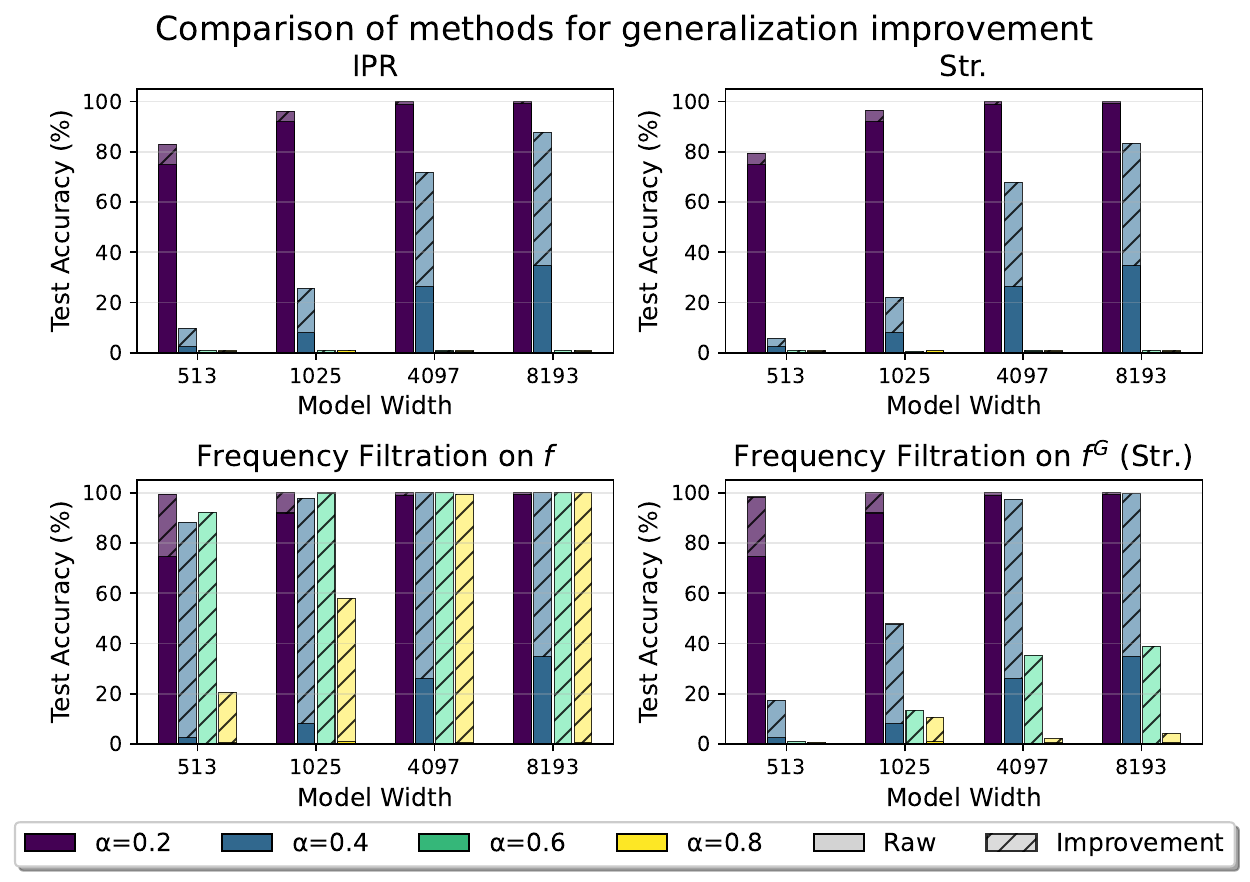}
  \caption{Generalization improvement on modular addition. IPR and Str. achieve similar performance. Applying FF to the sub-network $f^G$ selected by Str. further improves generalization, but the gain remains substantially smaller than applying FF directly to the full model $f$.}
  \label{fig:improvement-comparison}
\end{figure}

\begin{result}\label{res:limited-improvement}
Generalization improvement from neuron selection/pruning is limited compared with FF.
\end{result}

Figure~\ref{fig:improvement-comparison} shows that neuron selection can improve generalization, especially under light label noise, and that the task-agnostic Str. score performs comparably to the task-specific IPR score. However, this improvement is much weaker than the gain achieved by applying FF directly to the full model. Even applying FF after selecting the generalization sub-network $f^G$ does not recover the performance of FF on the original network. This contrast indicates that generalizable structure is not localized to a small subset of neurons; instead, it is diffusely encoded across the network and intertwined with components used for noise memorization.

This comparison clarifies the limitation of task-agnostic post-hoc neuron selection. Unlike FF, which uses the domain-specific Fourier structure of the task to extract the latent rule, neuron selection can only identify neurons that are more important on average. As a result, pruning may partially improve generalization but cannot fully unleash the rule already learned by the model. This suggests two broader implications: future learning paradigms may need to explicitly isolate memorization from logical inference during training, and post-hoc knowledge extraction may require domain-specific structural priors rather than generic neuron-importance scores.

\paragraph{Conclusion.} Our results demonstrate that over-parameterized models can internally learn underlying rules even under explicit label noise, though this generalization remains heavily entangled with memorization across distributed neurons. While task-agnostic neuron selection provides only limited improvement, frequency filtration can successfully recover latent rules when a suitable structural prior is available. Crucially, we observe that these core entanglement mechanisms extend beyond algebraic tasks to architectures like Transformers and datasets such as MNIST. We acknowledge that our setup inherently abstracts away the complex variability of natural vision or language tasks, but this controlled setting is necessary to provide a rigorous, verifiable boundary between structural generalization and unstructured memorization. Ultimately, this foundational study highlights a fundamental limitation of post-hoc pruning-based extraction, suggesting that fully disentangling these processes in practical applications will require novel training paradigms or architectural designs with explicit function separation.

\bibliography{refs}


\appendix

\section{Related Work}\label{sec:related-work}
\paragraph{Model-wise Double Descent and Over-parameterization.}
The discovery of the double descent phenomenon (illustrated in Figure \ref{fig:double_descent}) marked a significant shift in modern machine learning, suggesting that the classical bias–variance trade-off is incomplete \citep{belkin2018reconceiling,yang2020rethinking}. In the over-parameterized regime, test error can decrease again after the interpolation threshold, leading to superior generalization. Theoretically, this second descent has been rigorously characterized in linear regression under various assumptions \citep{belkin2020two, advani2020high, adlam2020neural,bartlett2020benign, mei2022generalization, hastie2022surprises,muthukumar2020harmless}, although some studies indicate that appropriate regularization can mitigate or eliminate this effect \citep{nakkiran2020optimal}. In non-linear settings, double descent has been robustly observed in computer vision tasks, particularly in the presence of label noise \citep{nakkiran2021deep,umar2025effect}.

Recent efforts have sought mechanistic explanations for the generalization power of large-scale models. One perspective proposes that over-parameterized models generalize through smoother interpolation of noisy data points \citep{gamba2022deep,somepalli2022can}. Furthermore, the emergence of neural scaling laws has been increasingly attributed to the superposition of discrete features or representations \citep{liu2025superposition}. \citet{davies2023unifying,huang2024unified} further unify double descent with grokking by framing both phenomena as a competition between memorization and generalization circuits. Our work extends this line of research by investigating how model size specifically governs the resolution of this competition under label noise.

\paragraph{Grokking and Emergent Generalization.}
In over-parameterized models, generalization often emerges long after the training loss has vanished, a phenomenon known as grokking \citep{power2022grokking}. This decoupling between fitting and rule discovery is prominently observed in modular arithmetic tasks across various architectures \citep{liu2022towards, nandaprogress,mallinaremergence}. While specific algorithms, such as Pizza and Clock, have been identified for Transformers solving these tasks \citep{nandaprogress, zhong2023clock}, and a universal abstract algorithm for deep neural networks has been uncovered \citep{mccracken2025uncovering}, the exact analytical solution for two-layer ReLU networks remains unsolved. Existing theoretical works have explored provable solutions using quadratic activations or specific regularization schemes \citep{gromov2023grokking, tian2025provable}, yet the role and impact of label noise in these settings remain largely unexplored.

\paragraph{Learning under Label Noise.}
Most existing work on learning with label noise focuses on robust training strategies to prevent overfitting \citep{li2020gradient,liu2022robust}. We refer readers to \citet{song2022learning} for a comprehensive survey. \citet{xu2023benign} theoretically analyze the performance of two-layer ReLU networks on noisy XOR data under the assumption of fixed second-layer weights, which deviates from standard end-to-end training dynamics. \citet{doshigrok} examine the effects of noise ratios and different regularization methods on arithmetic tasks. In contrast, we primarily focus on the relationship between model size and generalization performance, and further investigate the mechanistic entanglement between memorization and generalization.

\section{Analytical Solution Under Quadratic Activation}\label{sec:quadratic_proof}
The fact that the solution in Equations~\eqref{equ:exact-solution} is exact for quadratic activations has been well discussed in \citet{pearce2023grokking,morwanifeature}. 
We present their theory here, adapted to our notation, in Proposition~\ref{prop:analytical_solution}.
\begin{proposition}\label{prop:analytical_solution}
(Theorem 7, \citet{morwanifeature}) For the modular addition task, any two-layer network $f_{\theta} = \bm{W}\left(\bm{U} \bm{e}_a + \bm{V}\bm{e}_b\right)^2$ with $M\ge 4(P-1)$ that maximizes the margin on $\mathcal{D}_{\text{total}}$ satisfies the following conditions:

1. for each neuron $\left\{\bm{u}_m, \bm{v}_m, \bm{w}_m\right\}$, there exists a scaling constant $\lambda \in \mathbb{R}$ and a frequency $\omega \in \left\{1, \cdots, \frac{P-1}{2}\right\}$, such that
\begin{subequations}
\begin{align}
& u_{mi} = \lambda \cos \left( \frac{2\pi}{P}\omega_m i + \varphi^{(a)}_m \right), \\
& v_{mj} = \lambda \cos \left( \frac{2\pi}{P}\omega_m j + \varphi^{(b)}_m \right), \\
& w_{mk} = \lambda \cos \left( \frac{2\pi}{P}\omega_m k + \varphi^{(c)}_m \right), 
\end{align}
\label{equ:exact-solution-appendix}
\end{subequations}
for some phase offsets $\varphi^{(a)}_m, \varphi^{(b)}_m, \varphi^{(c)}_m \in \mathbb{R}$ satisfying $\varphi^{(a)}_m + \varphi^{(b)}_m = \varphi^{(c)}_m$. 

2. For every frequency $\omega \in \left\{1, \cdots, \frac{P-1}{2}\right\}$, at least one neuron in the network uses this frequency.
\end{proposition}

\section{Supplementary Experiments for Section \ref{sec:double-descent}}\label{sec:additional-exp-main-results}
\subsection{Double-descent curves on varying setups}\label{sec:addition-robustly-double-descent}
Model-wise double descent is clearly observed across different activation functions, training ratios, optimizers, and arithmetic tasks. We summarize the Figures about it in Table \ref{tab:figure_indices_double_descent}.
\begin{table}[htbp]
\centering
\caption{Figure indices showing double descent curves under varying setups.}
\label{tab:figure_indices_double_descent}
\begin{tabular}{|c|c|c|c|}
\hline
Activation Functions & Training Ratios & Optimizers & Arithmetic Tasks \\
\hline
\ref{fig:across_activations}, \ref{fig:quadratic_across_wds} & \ref{fig:train_ratio_0.4} (40\%), \ref{fig:AdamW_wds}(50\%), \ref{fig:train_ratio_0.6}(60\%) & \ref{fig:AdamW_wds} (AdamW), \ref{fig:Adam_wds} (Adam), \ref{fig:Muon_wds} (Muon) & \ref{fig:across_tasks} \\
\hline
\end{tabular}
\end{table}


\begin{figure*}[!htbp] 
  \centering
  \includegraphics[width=0.6\linewidth]{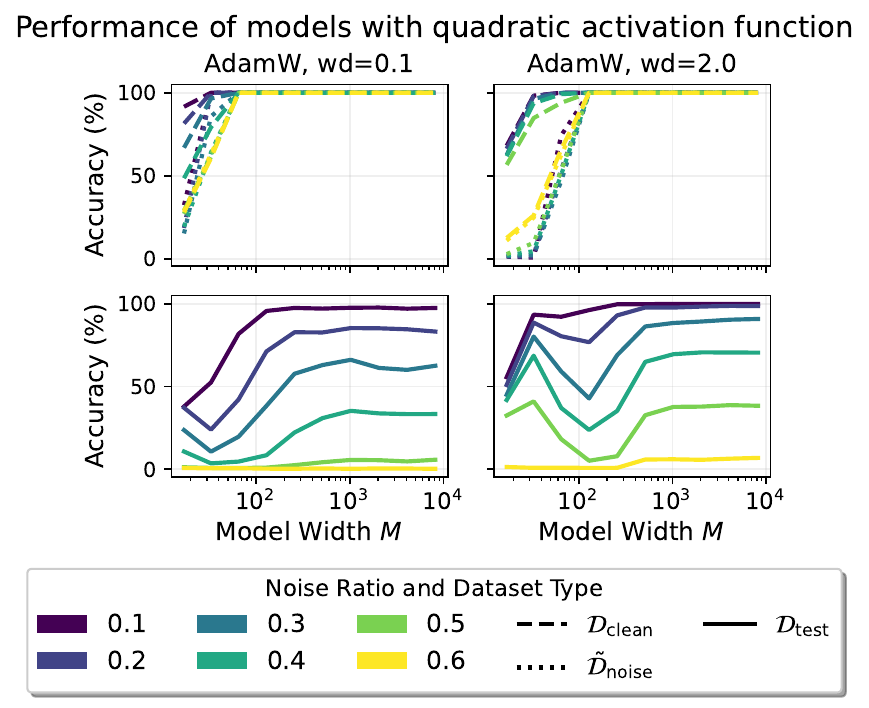}
  \caption{Ceilings of test accuracy for quadratic activation functions exist across varying weight decays.}
  \label{fig:quadratic_across_wds}
\end{figure*}

\begin{figure*}[!htbp] 
  \centering
  \includegraphics[width=0.7\linewidth]{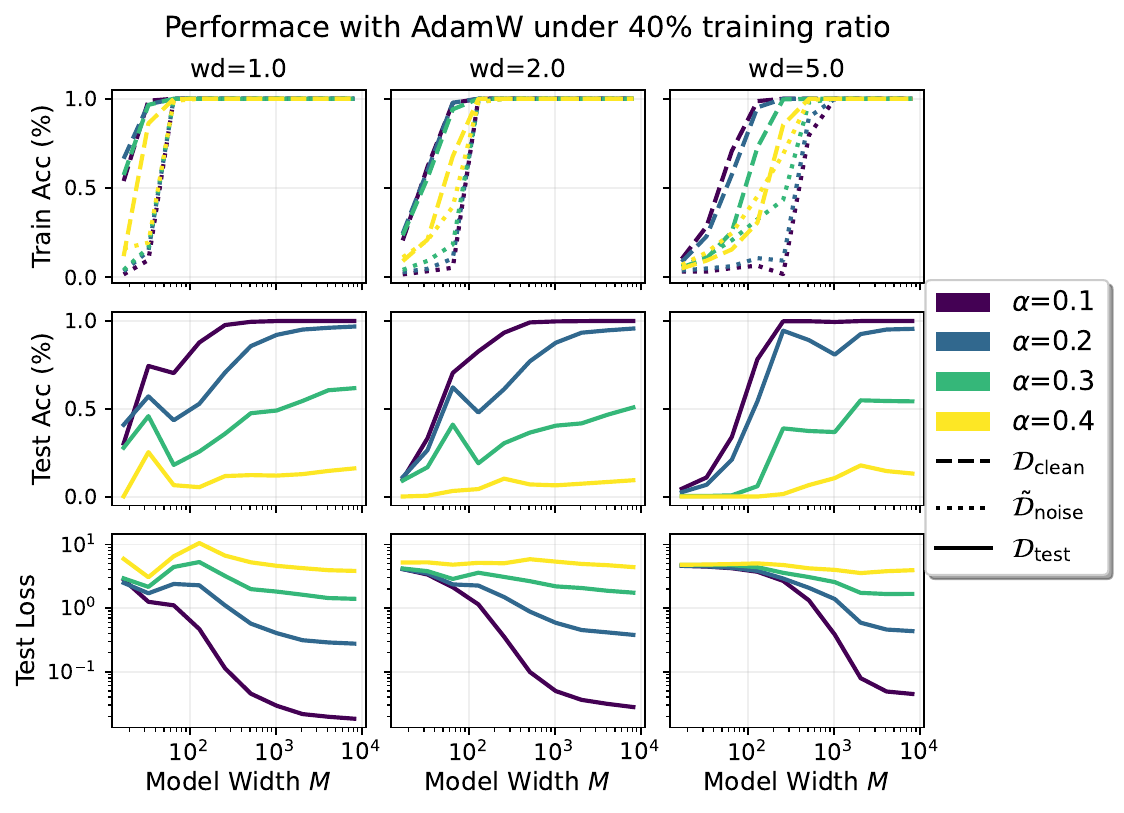}
  \caption{Performance of AdamW under 40\% training ratio.}
  \label{fig:train_ratio_0.4}
\end{figure*}

\begin{figure*}[!htbp] 
  \centering
  \includegraphics[width=0.7\linewidth]{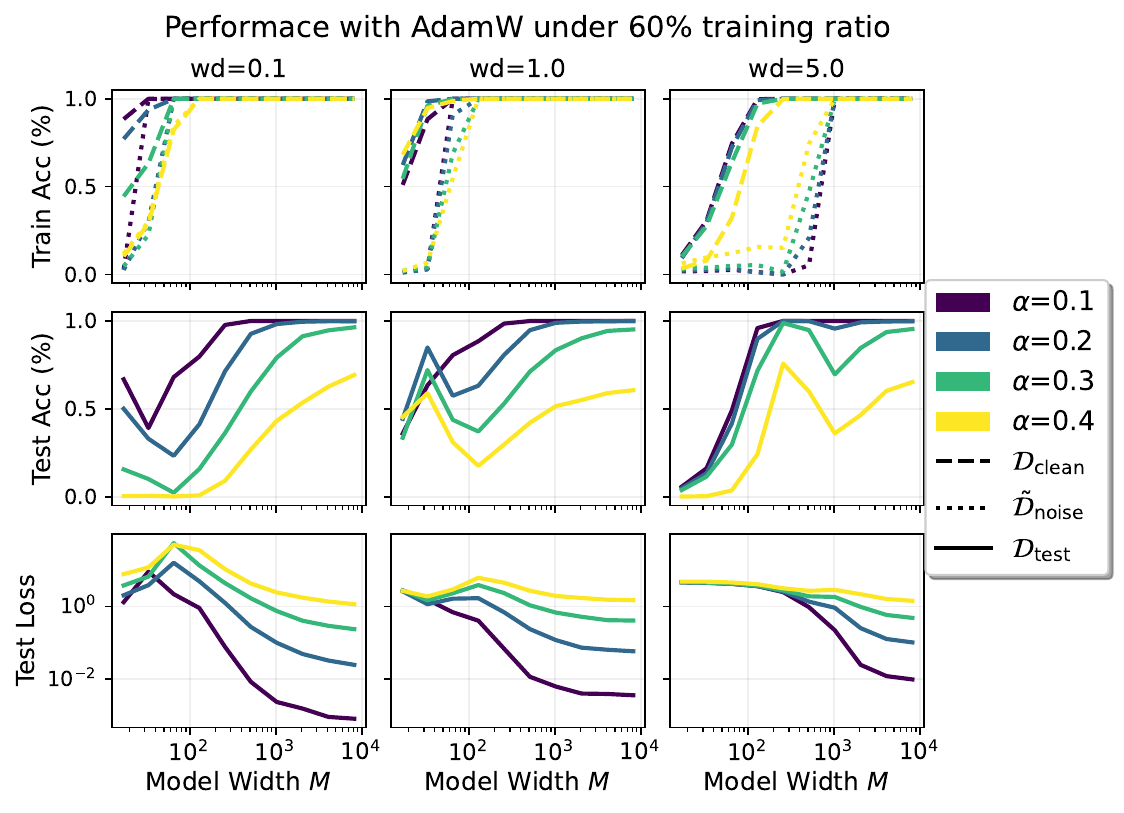}
  \caption{Performance of AdamW under 60\% training ratio.}
  \label{fig:train_ratio_0.6}
\end{figure*}

\begin{figure*}[!htbp] 
  \centering
  \includegraphics[width=0.9\linewidth]{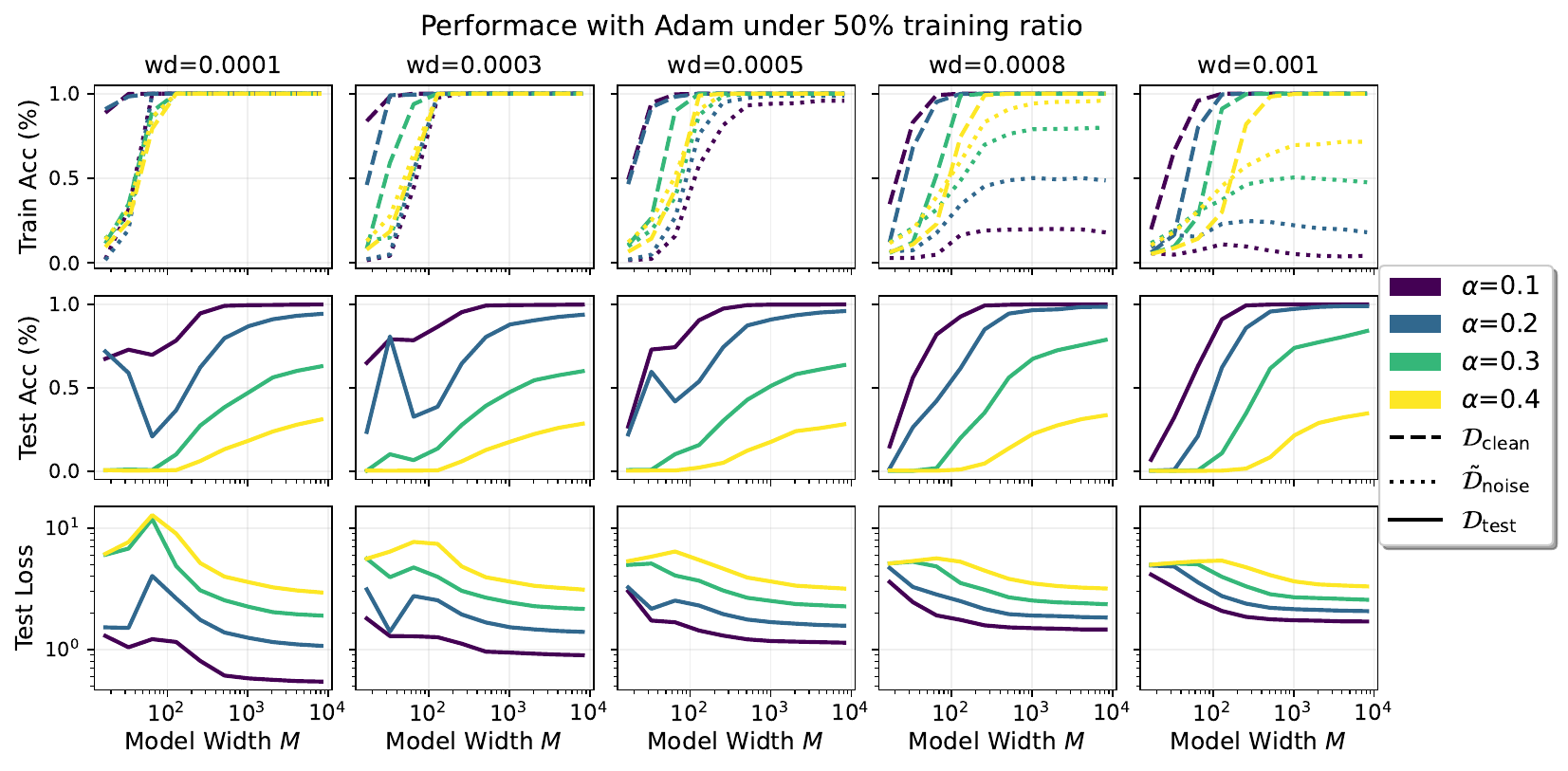}
  \caption{Performance of Adam under varying weight decays.}
  \label{fig:Adam_wds}
\end{figure*}

\begin{figure*}[!htbp] 
  \centering
  \includegraphics[width=0.95\linewidth]{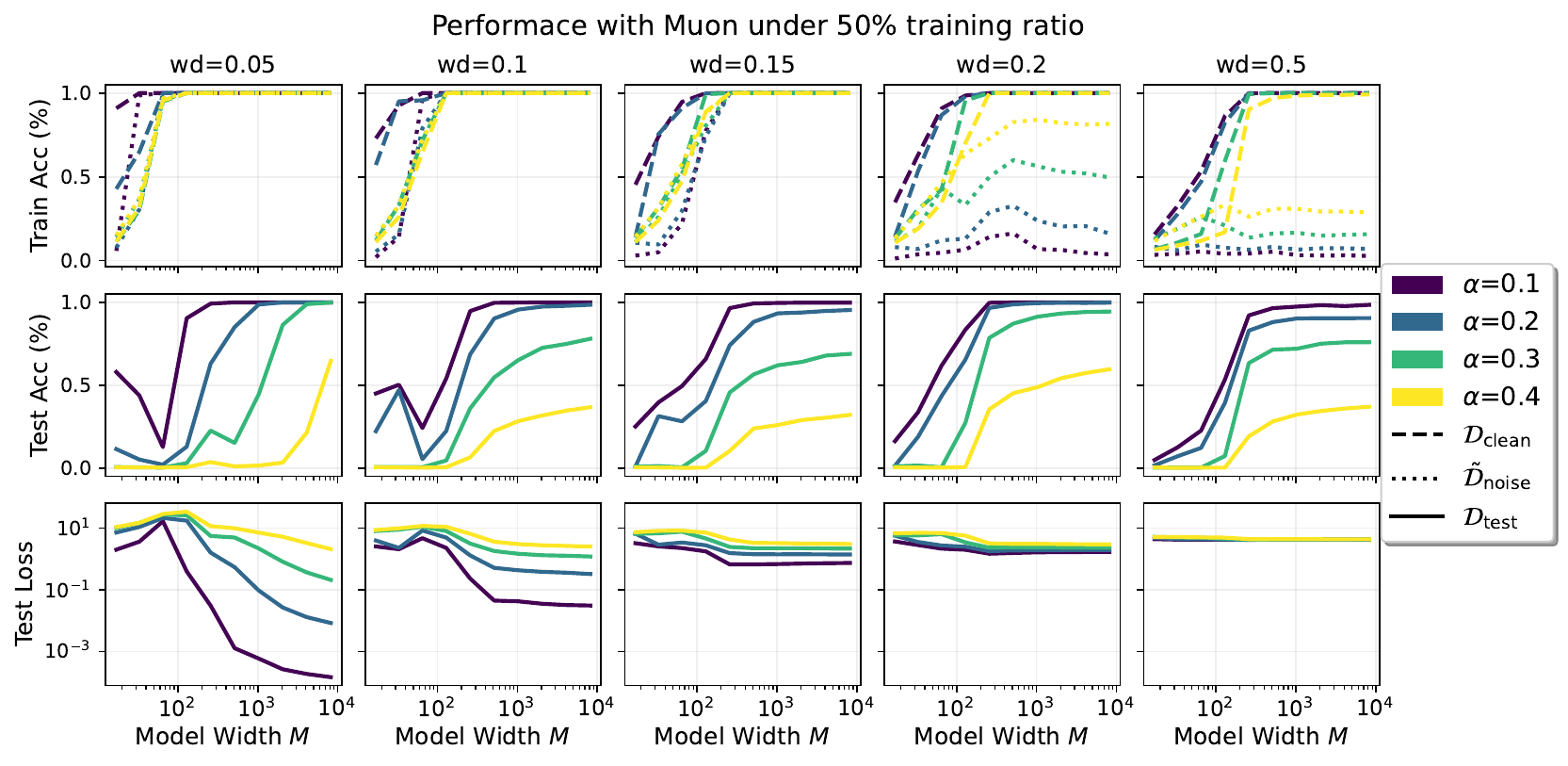}
  \caption{Performance of Muon under varying weight decays.}
  \label{fig:Muon_wds}
\end{figure*}

\begin{figure}[!htbp] 
  \centering
  \includegraphics[width=0.6\linewidth]{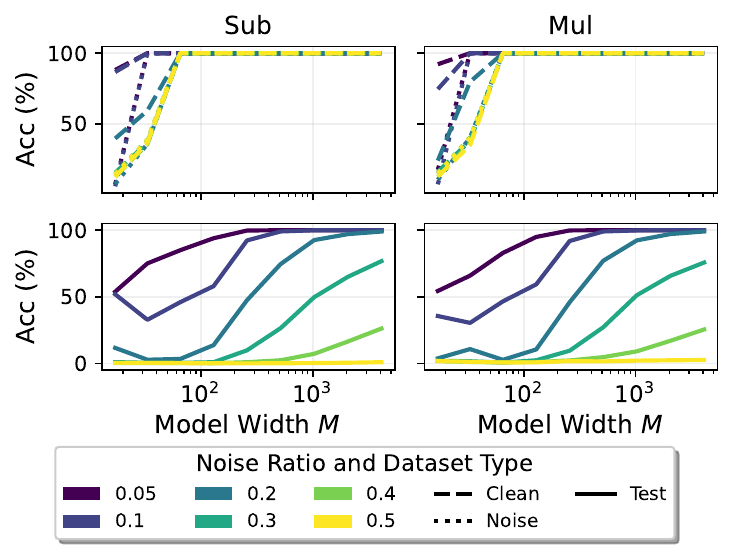}
  \caption{Performance on modular subtraction (\textit{left}) and multiplication (\textit{right}) tasks. Models are trained using AdamW under 50\% training ratio.}
  \label{fig:across_tasks}
\end{figure}
\begin{figure*}[!htbp] 
  \centering
  \includegraphics[width=0.9\linewidth]{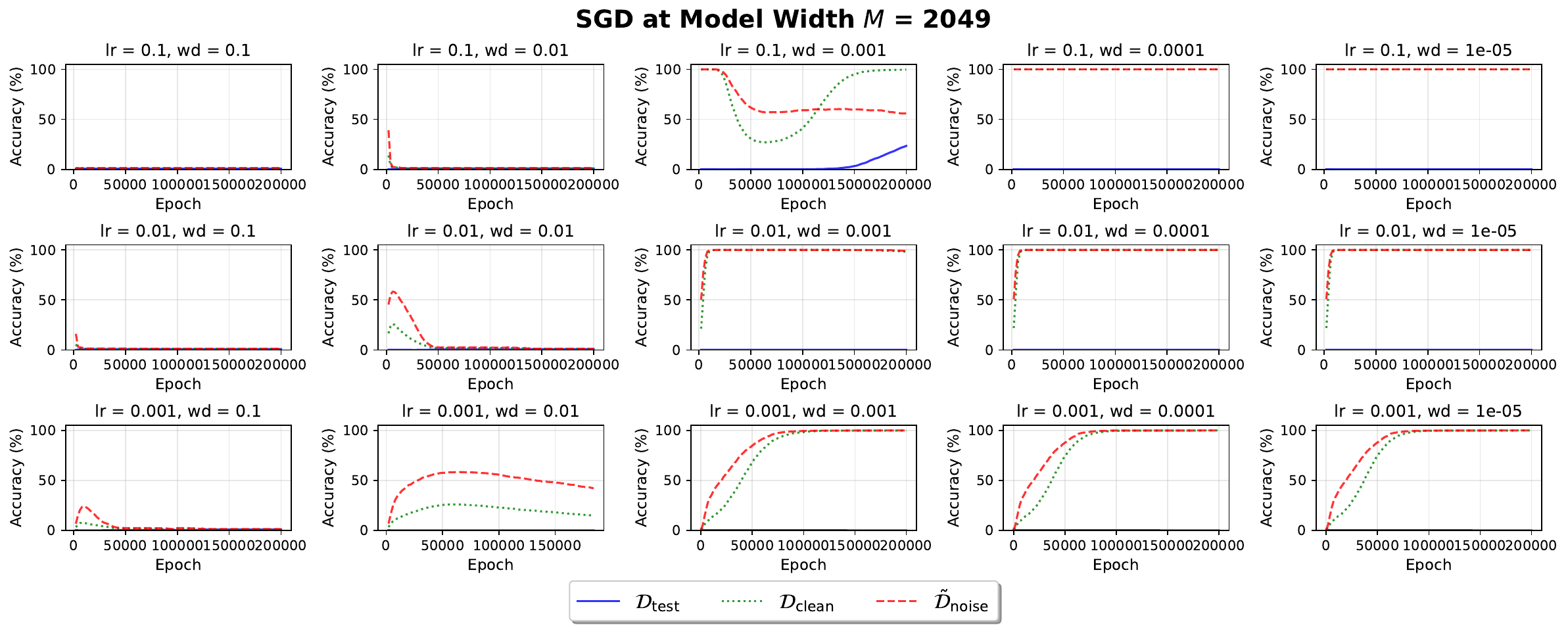}
  \caption{Training dynamics of SGD under different learning rates (lr) and weight decay (wd) values at $M=2049$ and $\alpha=0.3$. Results are reported every 2{,}000 epochs, where the total epochs are 200{,}000. }
  \label{fig:SGD_dynamics}
\end{figure*}

\paragraph{SGD is not a good choice.}
Figure~\ref{fig:SGD_dynamics} shows the training dynamics of SGD under different choices of learning rate and weight decay. Among all configurations, only the setting with $\text{lr}=0.1$ and $\text{wd}=0.001$ exhibits noticeable generalization behavior. This suggests that SGD requires highly restrictive hyperparameter tuning to achieve reasonable generalization. Moreover, even at epoch 200{,}000, the test accuracy remains relatively low and is still increasing, indicating that SGD converges extremely slowly. Overall, SGD is inefficient for this task, as it consumes an excessive number of training epochs to reach a modest level of generalization.

\subsection{Discussion on Model Misspecification}\label{sec:model-misspecification}

\begin{figure}[!htbp] 
  \centering
  \includegraphics[width=0.8\linewidth]{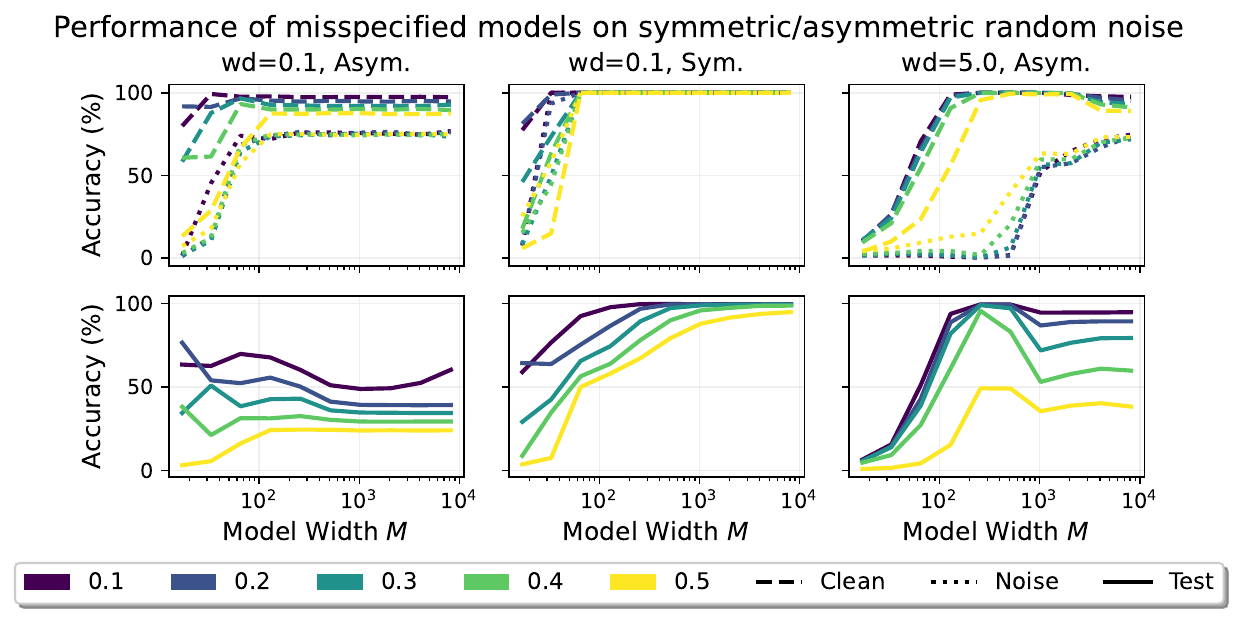}
  \caption{Performance of first-layer tied models under model misspecification. 
\textit{(Left)} With asymmetric random noise, the tied model fails to fully interpolate noisy labels and exhibits irregular generalization behavior. 
\textit{(Middle)} Under symmetric noise, the tied model successfully interpolates all noisy labels, and generalization improves with increasing model size. 
\textit{(Right)} Stronger regularization mitigates the adverse effect of misspecification by encouraging the model to ignore noisy labels.
}
  \label{fig:tied_model}
\end{figure}

We consider a scenario where the model architecture is not expressive enough to memorize all noisy labels, which leads to atypical generalization behavior.

Modular addition and multiplication are commutative operations, i.e., $a \circ b = b \circ a$. Accordingly, an alternative architecture enforces a tied first layer by setting $U = V$ \citep{pearce2023grokking}. Under random label noise, however, the noisy samples generally violate this symmetry. As a result, the tied model is unable to interpolate all noisy labels, and the coexistence phase does not emerge in this setting, leading to irregular test accuracy curves (see the left panel in Figure~\ref{fig:tied_model}).

To verify that the asymmetric structure of random noise is responsible for this behavior, we construct a symmetric noise pattern and re-evaluate the tied model. Specifically, for a noise ratio $\alpha$ in modular addition, we randomly sample $\frac{\alpha}{2}|\mathcal{D}_{\text{train}}|$ examples to form $\mathcal{D}_{\text{asym}}$, where each $(a,b,c)\in\mathcal{D}_{\text{asym}}$ satisfies $a<b$. The noisy dataset is then constructed as
\[
\tilde{\mathcal{D}}_{\text{noise}}=\{(a,b,\tilde{c}) \mid (a,b,c)\in\mathcal{D}_{\text{asym}} \text{ or } (b,a,c)\in\mathcal{D}_{\text{asym}}, \ \tilde{c}\neq c,\ \tilde{c}\in\{0,\ldots,P-1\}\}.
\]
Under this symmetric noise, the tied model exhibits predictable behavior: noisy labels can be fully memorized once the model width exceeds a threshold (i.e., the coexistence phase appears), and generalization improves with increasing model size (see the middle panel in Figure~\ref{fig:tied_model}).

When the noise structure is unknown, model misspecification becomes a potential risk. One practical mitigation strategy is to strengthen regularization so as to extend the inversion phase. The tied model is able to interpolate all clean samples, and increasing weight decay encourages the model to suppress/neglect noisy labels, thereby recovering good generalization performance (see the right panel in Figure~\ref{fig:tied_model}).

\subsection{Discussion on Result \ref{res:noise-memorize-faster}}\label{sec:discussion-noise-memorize-faster}

We provide a gradient-based explanation for the counter-intuitive observation in
Result~\ref{res:noise-memorize-faster}. Let
$\mathcal{D}_{\mathrm{c}}=\mathcal{D}_{\text{clean}}$ and
$\mathcal{D}_{\mathrm{n}}=\tilde{\mathcal{D}}_{\text{noise}}$ denote the clean and
noisy subsets of the training data, respectively. For a model parameterized by
$\theta$, we decompose the empirical loss as
\[
    L(\theta)
    =
    L_{\mathrm{c}}(\theta) + L_{\mathrm{n}}(\theta),
    \qquad
    L_{\mathrm{c}}(\theta)
    =
    \sum_{z\in\mathcal{D}_{\mathrm{c}}}\ell(z;\theta),
    \qquad
    L_{\mathrm{n}}(\theta)
    =
    \sum_{z\in\mathcal{D}_{\mathrm{n}}}\ell(z;\theta),
\]
where $\ell$ is the cross-entropy loss on a single example. Under full-batch
gradient descent, the update direction is
\[
    -\nabla L(\theta)
    =
    -\nabla L_{\mathrm{c}}(\theta)
    -
    \nabla L_{\mathrm{n}}(\theta).
\]
For a sufficiently small step size $\eta$, the first-order decrease of the clean
and noisy losses after one update is approximately
\[
    \Delta L_{\mathrm{c}}
    \approx
    -\eta
    \left\langle
        \nabla L_{\mathrm{c}},
        \nabla L_{\mathrm{c}}+\nabla L_{\mathrm{n}}
    \right\rangle,
    \qquad
    \Delta L_{\mathrm{n}}
    \approx
    -\eta
    \left\langle
        \nabla L_{\mathrm{n}},
        \nabla L_{\mathrm{c}}+\nabla L_{\mathrm{n}}
    \right\rangle .
\]
Thus, the instantaneous learning speed of each subset is controlled by how well
its own gradient aligns with the full-batch training direction. To compare the
two subsets fairly, we normalize by their sizes and define the per-sample loss
decrease rates
\[
    s_{\mathrm{c}}
    =
    \frac{
    \left\langle
        \nabla L_{\mathrm{c}},
        \nabla L_{\mathrm{c}}+\nabla L_{\mathrm{n}}
    \right\rangle
    }{|\mathcal{D}_{\mathrm{c}}|},
    \qquad
    s_{\mathrm{n}}
    =
    \frac{
    \left\langle
        \nabla L_{\mathrm{n}},
        \nabla L_{\mathrm{c}}+\nabla L_{\mathrm{n}}
    \right\rangle
    }{|\mathcal{D}_{\mathrm{n}}|}.
\]
Empirically, at the beginning of training, we observe that
$\langle \nabla L_{\mathrm{c}}, \nabla L_{\mathrm{n}}\rangle < 0$, meaning that
the clean and noisy subsets induce partially conflicting gradient directions.
Nevertheless, both normalized progress measures remain positive, i.e.,
$s_{\mathrm{c}}>0$ and $s_{\mathrm{n}}>0$. This indicates that the full-batch
update still decreases both the clean and noisy losses at the very beginning of
training, although it does so at different per-sample rates. However, since the
noisy subset is much smaller when the noise ratio $\alpha$ is small, the update
direction can produce a larger \emph{per-sample} decrease on the noisy subset.
In other words, although the noisy labels are unstructured, each noisy example
receives a stronger individual push from the full-batch gradient. This explains
why the noisy-label accuracy can increase earlier than the clean-label accuracy,
especially at lower noise ratios.

This gradient-based explanation is also consistent with additional experiments beyond modular arithmetic. To test whether the early noisy-over-clean learning order is purely induced by the algebraic structure of modular arithmetic, we further train a three-layer neural network on MNIST with randomly corrupted labels(hidden dimensions$=[1024, 256]$, using AdamW with learning rate $0.001$). As shown in Table~\ref{tab:mnist-noisy-faster}, under full-batch training, noisy-label accuracy is higher than clean-label accuracy at epoch 1 for both $\alpha=0.05$ and $\alpha=0.2$. In the next epoch, clean accuracy quickly catches up and becomes much larger, while noisy accuracy drops. These results suggest that the early advantage of noisy labels is not solely a consequence of the Fourier structure of modular arithmetic, but can also arise in a standard image classification task under a similar optimization protocol.

\begin{table}[t]
\centering
\caption{Early-stage clean and noisy accuracy on MNIST with a three-layer neural
network under full-batch training.}
\label{tab:mnist-noisy-faster}
\begin{tabular}{lcccc}
\toprule
 & \multicolumn{2}{c}{$\alpha=0.05$}
 & \multicolumn{2}{c}{$\alpha=0.2$} \\
\cmidrule(lr){2-3}\cmidrule(lr){4-5}
Period
& Noisy acc. & Clean acc.
& Noisy acc. & Clean acc. \\
\midrule
Epoch 1 & $10.87\%$ & $2.86\%$  & $10.75\%$ & $2.89\%$  \\
Epoch 2 & $6.47\%$  & $46.29\%$ & $5.71\%$  & $48.09\%$ \\
\bottomrule
\end{tabular}
\end{table}

The MNIST experiment also helps clarify the scope of the phenomenon. The
effect is most visible at the very beginning of full-batch training and should
not be interpreted as saying that noisy labels are ultimately easier to learn
than clean labels. Instead, the first few updates may create a larger
per-sample progress on the smaller noisy subset, while the clean structure
dominates soon afterward. This is consistent with our gradient analysis: the
initial quantities $s_{\mathrm{c}}$ and $s_{\mathrm{n}}$ can both be positive,
with $s_{\mathrm{n}}>s_{\mathrm{c}}$, even though the clean and noisy gradients
partially conflict.

This explanation also clarifies why the phenomenon depends on the training
protocol. Under full-batch training, the gradient contribution of the noisy
subset is accumulated coherently in every update, so the direction that benefits
the current noisy labels is repeatedly applied. In contrast, under small-batch
training, different mini-batches contain different noisy examples, and the
gradient direction that helps memorize the noisy labels in one batch is generally
not transferable to another batch. As a result, the per-sample push toward any
specific corrupted label becomes less stable, and the early noisy-over-clean
learning order is weakened or may disappear.

\subsection{Accuracy learning curves}\label{sec:additional-training-dynamics}
The accuracy learning curves capture when memorization and generalization emerge during training. 
We observe the grokking phenomenon even under noisy labels: the model quickly reaches 100\% training accuracy, while test accuracy improves significantly only later. 
To illustrate this, we separately present the training dynamics on short and long time scales, highlighting memorization and generalization, respectively.

\subsubsection{Memorization on a short time scale}
\begin{figure*}[!htbp] 
  \centering
  \includegraphics[width=0.95\linewidth]{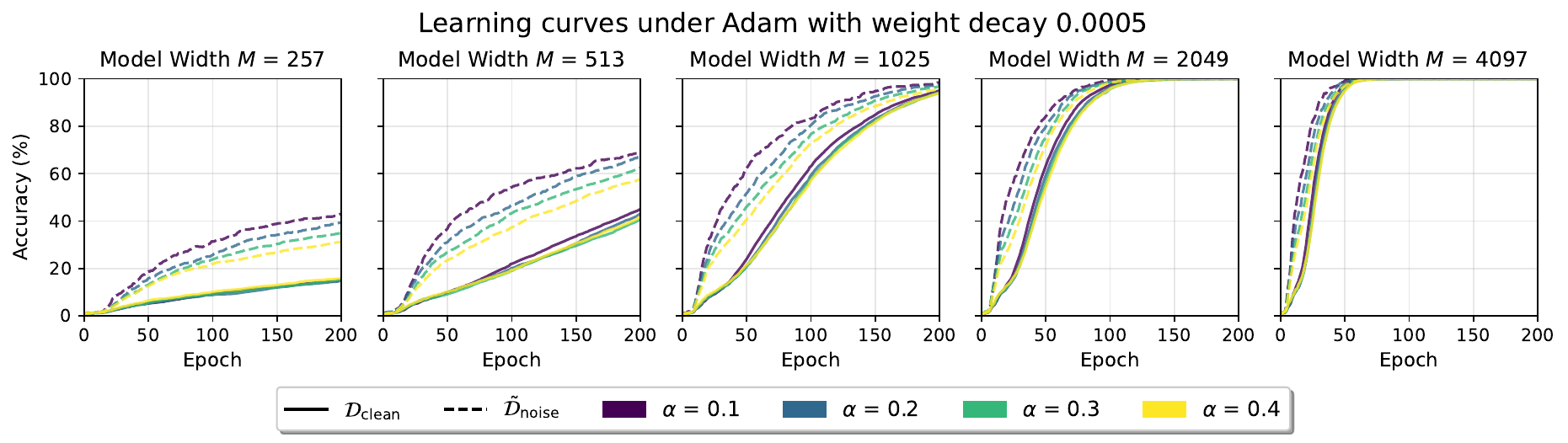}
  \caption{Memorization curves for the Adam optimizer. Larger models reach high training accuracy more quickly. Noisy labels are memorized faster than clean labels, and lower noise ratios further accelerate this effect.}
  \label{fig:training_dynamics_short_Adam}
\end{figure*}

\begin{figure*}[!htbp] 
  \centering
  \includegraphics[width=0.95\linewidth]{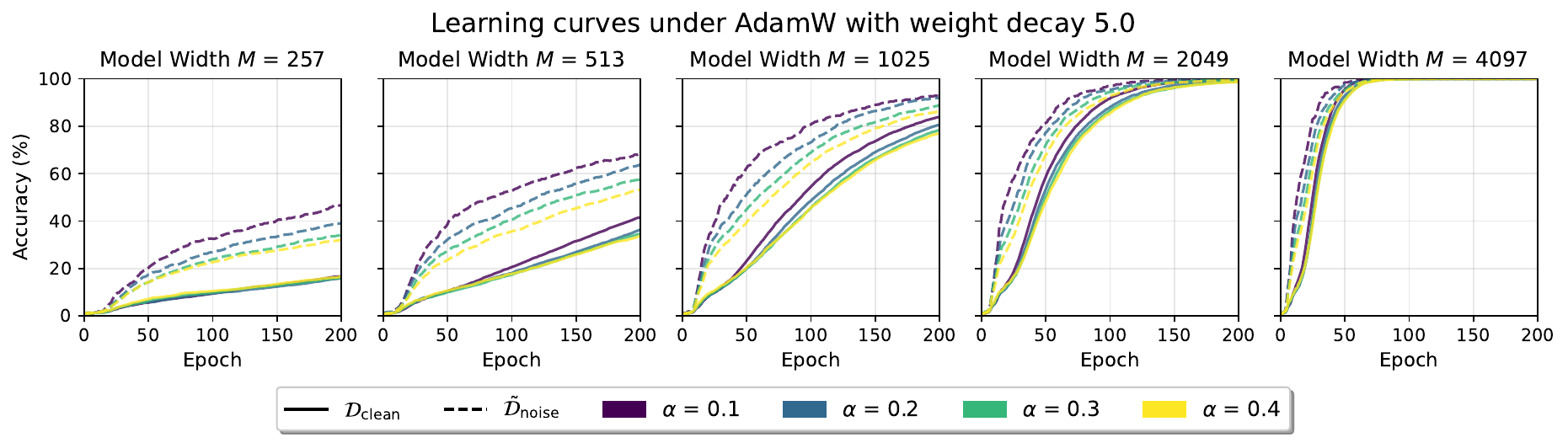}
  \caption{Memorization curves for AdamW optimizer.}
  \label{fig:training_dynamics_short_AdamW}
\end{figure*}

\begin{figure*}[!htbp] 
  \centering
  \includegraphics[width=0.95\linewidth]{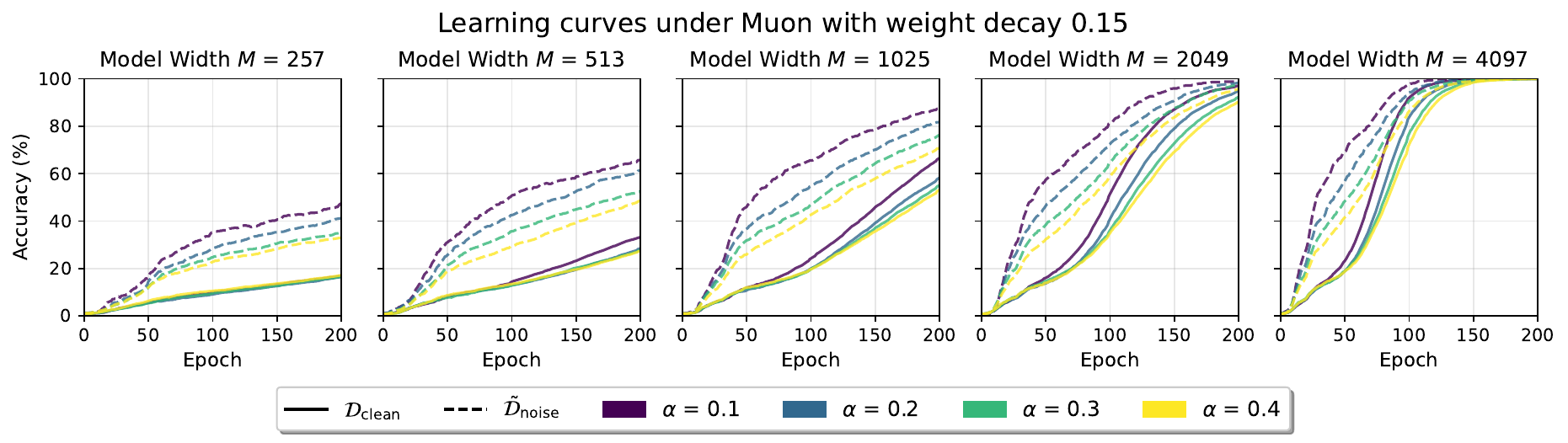}
  \caption{Memorization curves for Muon optimizer.}
  \label{fig:training_dynamics_short_Muon}
\end{figure*}

Without noise in the training dataset, previous studies have found that larger models memorize the training data faster \citep{tirumala2022memorization,huang2024unified}. 
This trend persists under noisy training data, although memorization speeds differ for clean and noisy subsets. 
We find that the model memorizes noisy labels faster than clean labels, and that the gap is larger at lower noise ratios. 
These phenomena are consistently observed across several optimizers, including Adam (Figure~\ref{fig:training_dynamics_short_Adam}), AdamW (Figure~\ref{fig:training_dynamics_short_AdamW}), and Muon (Figure~\ref{fig:training_dynamics_short_Muon}).

\subsubsection{Generalization on a long time scale}
\begin{figure*}[!htbp] 
  \centering
  \includegraphics[width=0.9\linewidth]{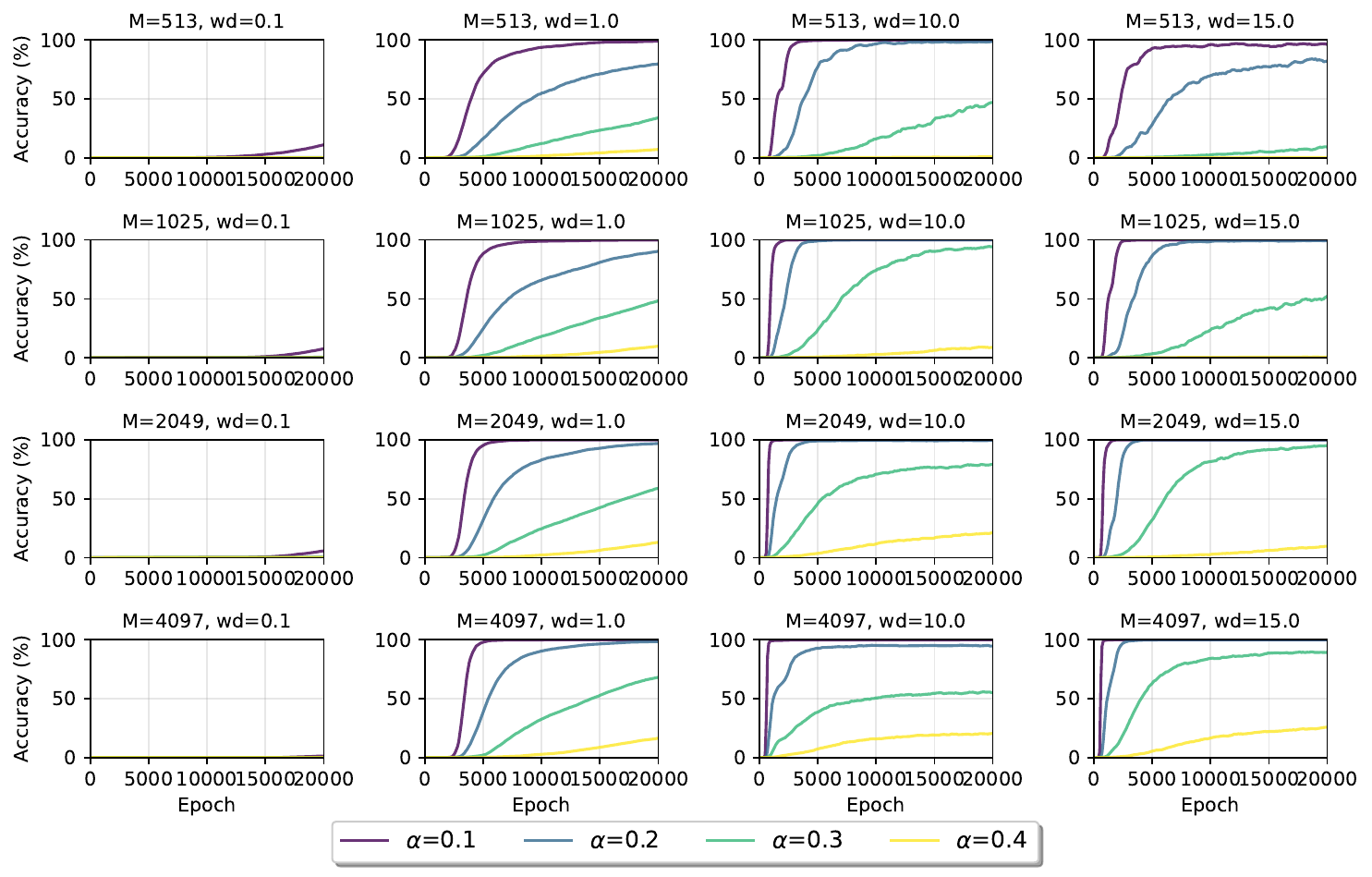}
  \caption{Training dynamics of test accuracy across various weight decays for AdamW. Small weight decay (wd=0.1) leads to delayed and slow grokking.}
  \label{fig:training_dynamics_M_vs_wd_AdamW}
\end{figure*}

\begin{figure*}[!htbp] 
  \centering
  \includegraphics[width=0.9\linewidth]{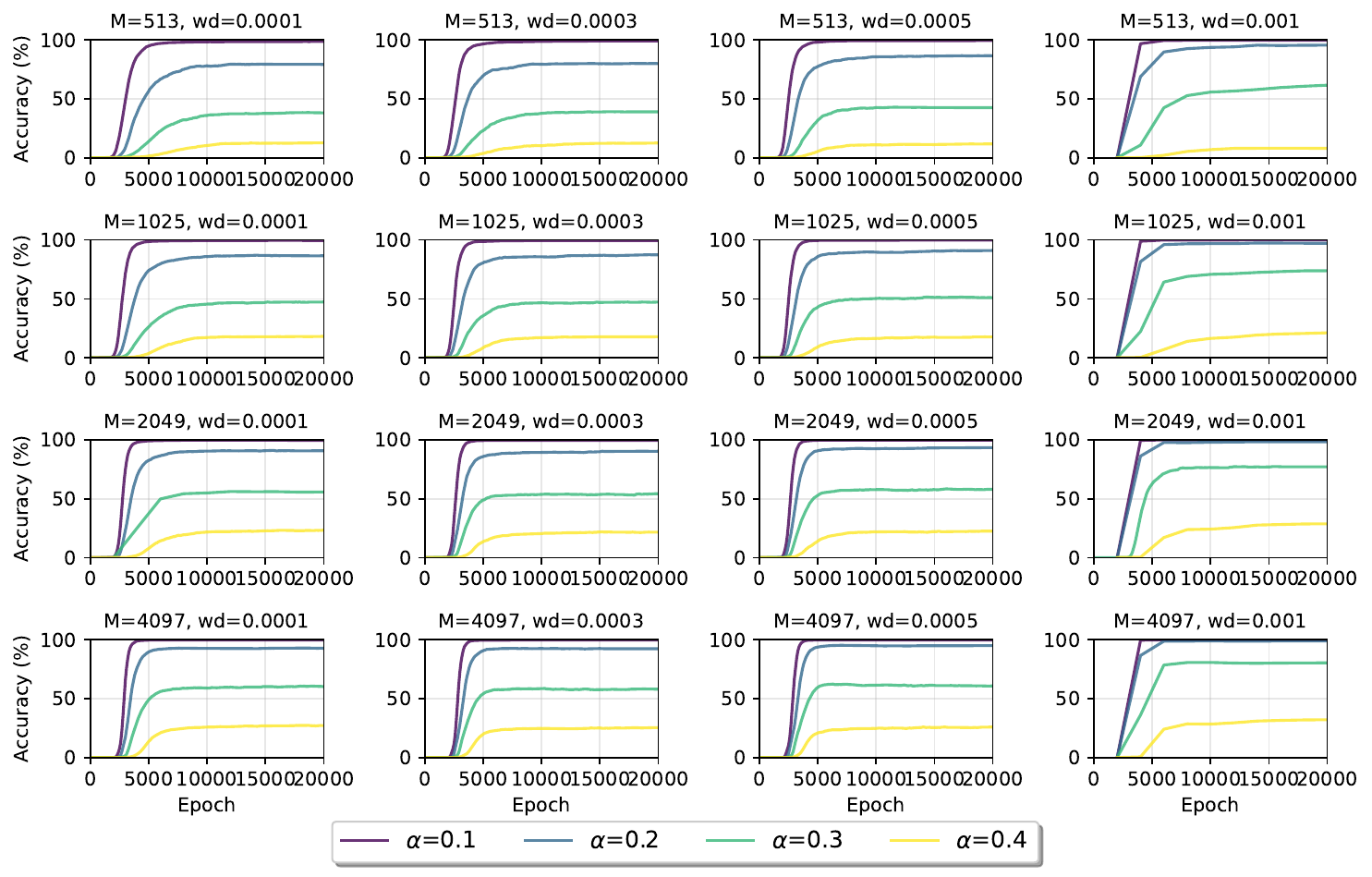}
  \caption{Training dynamics of test accuracy across various weight decays for Adam.}
  \label{fig:training_dynamics_M_vs_wd_Adam}
\end{figure*}

\begin{figure*}[!htbp] 
  \centering
  \includegraphics[width=0.9\linewidth]{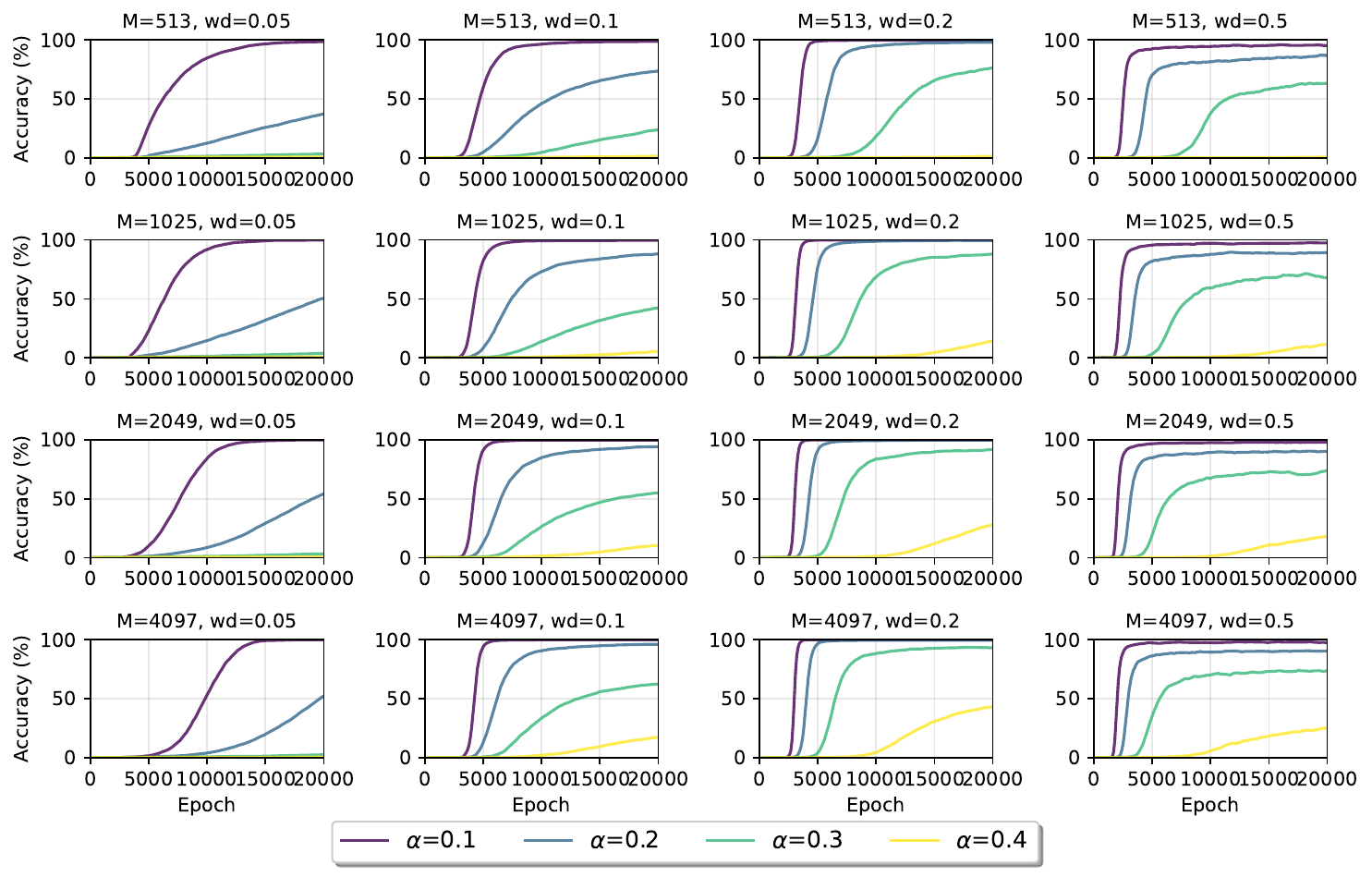}
  \caption{Training dynamics of test accuracy across various weight decays for Muon.}
  \label{fig:training_dynamics_M_vs_wd_Muon}
\end{figure*}

In terms of test accuracy, we observe a general tendency that models under larger weight decay grok faster, although the trend is not equally pronounced across all hyperparameter settings. This trend is clear for AdamW (Figure~\ref{fig:training_dynamics_M_vs_wd_AdamW}) and Muon (Figure~\ref{fig:training_dynamics_M_vs_wd_Muon}), but not evident for Adam (Figure~\ref{fig:training_dynamics_M_vs_wd_Adam}). Across all of them, models under smaller noise ratio grok faster.

\section{Supplementary Materials for Section \ref{sec:representation-of-generalization}}
\subsection{Details of frequency filtration}\label{sec:frequency_filtration_details}
In this section, we provide more details about the frequency filtration. That is to say, the details about decomposing $\left( \bm{u}_m, \bm{v}_m, \bm{w}_m \right)$, the $m^{\text{th}}$ neuron, into $\left( \bm{u}^G_m, \bm{v}^G_m, \bm{w}^G_m \right)$ and $\left( \bm{u}^R_m, \bm{v}^R_m, \bm{w}^R_m \right)$.

Let $\omega=e^{-2\pi i/P}$. Define the Fourier transform matrix $F\in \mathbb{C}^{P\times P}$ by
\begin{equation*}
F_{k,n} = \frac{1}{\sqrt{P}}\omega^{kn}=\frac{1}{\sqrt{P}}e^{-2\pi ikn/P}.
\end{equation*}
For any vector $\bm{x}\in\mathbb{R}^{P\times 1}$, denote its Fourier transform as
$
    \tilde{\bm{x}} = \bm{F} \bm{x}
$. $\bm{F}$ is a unitary matrix, so its conjugate transpose of $\bm{F}$, denoted as $\bm{F}^H$, gives $\bm{F}^H\bm{F}\bm{v} = \bm{v}$. Without loss of generality, we assume $\bm{x}$ is a vector satisfying that $\argmax_{k} \lvert\tilde{x}_k\rvert$ is a singleton.

We define the main frequency retrieval operator (for generalization part) as $\bm{F}^H \bm{F}_G(\bm{x})$, filtering the frequency with maximum magnitude, namely,
\begin{equation*}
    F_G(\bm{x})_{k,n} = \left\{
    \begin{array}{ll}
    F_{k,n}, & k \in \argmax_{k'} \lvert \tilde{x}_{k'} \rvert \\
    0, & k \notin \argmax_{k'} \lvert \tilde{x}_{k'} \rvert
    \end{array}
    \right.,
\end{equation*}
and the remaining frequency retrieval operator (for memorization) as $\bm{F}^H \bm{F}_R(\bm{x})$, where $\bm{F}_R(\bm{x})=\bm{F} - \bm{F}_G(\bm{x})$.

\begin{figure}[!htbp] 
  \centering
  \includegraphics[width=0.6\linewidth]{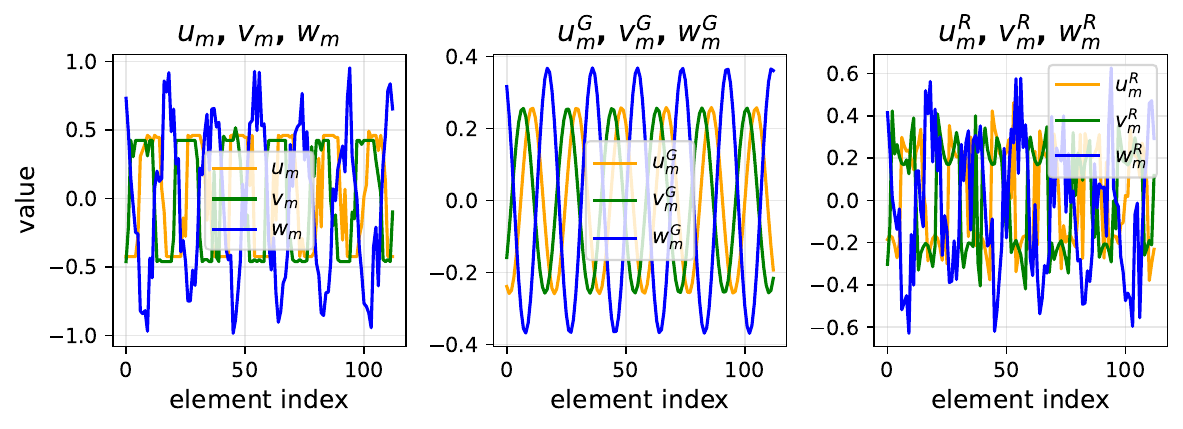}
  \caption{Visualization of applying frequency filtration to a neuron. $(\bm{u}^G_m, \bm{v}^G_m, \bm{w}^G_m)$ retains the frequency component with the maximum magnitude in $(\bm{u}_m, \bm{v}_m, \bm{w}_m)$.
}
  \label{fig:freq_filtration_on_neuron}
\end{figure}

For the $m^{\text{th}}$ neuron, let $\bm{P}_{G,m}=\bm{F}^H \bm{F}_G(\bm{w}_m)$ and $\bm{P}_{R,m}=\bm{F}^H \bm{F}_R(\bm{w}_m)$. Then, we define the decomposition to neuron $m$ as
\begin{align*}
 \bm{u}^G_m = \text{Real}(\bm{P}_{G,m} \bm{u}_m),\ \bm{u}^R_m = \text{Real}(\bm{P}_{R,m} \bm{u}_m), \\
 \bm{v}^G_m = \text{Real}(\bm{P}_{G,m} \bm{v}_m),\ \bm{v}^R_m = \text{Real}(\bm{P}_{R,m} \bm{v}_m), \\
 \bm{w}^G_m = \text{Real}(\bm{P}_{G,m} \bm{w}_m),\ \bm{w}^R_m = \text{Real}(\bm{P}_{R,m} \bm{w}_m). 
\end{align*}
It is easy to check $\bm{P}_{G,m} + \bm{P}_{R,m} = \bm{I}$, so 
\begin{equation*}
\left( \bm{u}_m, \bm{v}_m, \bm{w}_m \right) = \left( \bm{u}^G_m, \bm{v}^G_m, \bm{w}^G_m \right) + \left( \bm{u}^R_m, \bm{v}^R_m, \bm{w}^R_m \right).
\end{equation*}
We provide the visualization of a neuron under such frequency filtration in Figure \ref{fig:freq_filtration_on_neuron}.

\subsection{Frequency filtration benefits generalization under varying setups}\label{sec:freq-filtration-additional}

\begin{figure*}[!htbp] 
  \centering
  \includegraphics[width=0.6\linewidth]{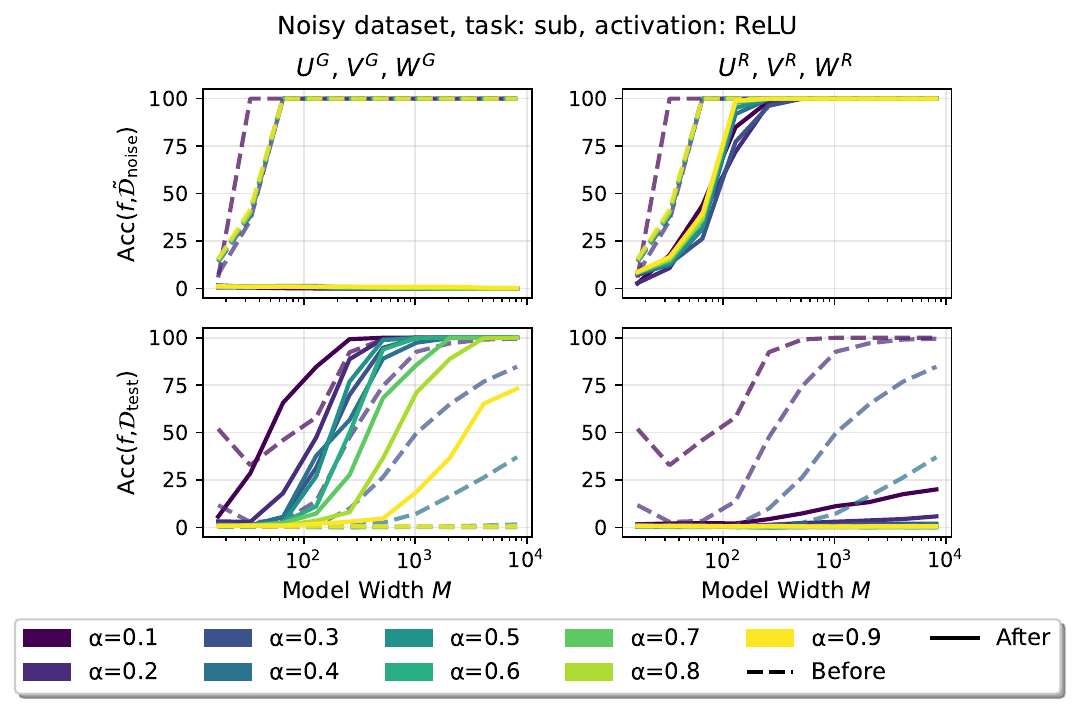}
  \caption{Performance before (dashed) and after (solid) the frequency filtration on the modular \textit{subtraction} task.}
  \label{fig:freq_filtration_sub}
\end{figure*}

\begin{figure*}[!htbp] 
  \centering
  \includegraphics[width=0.6\linewidth]{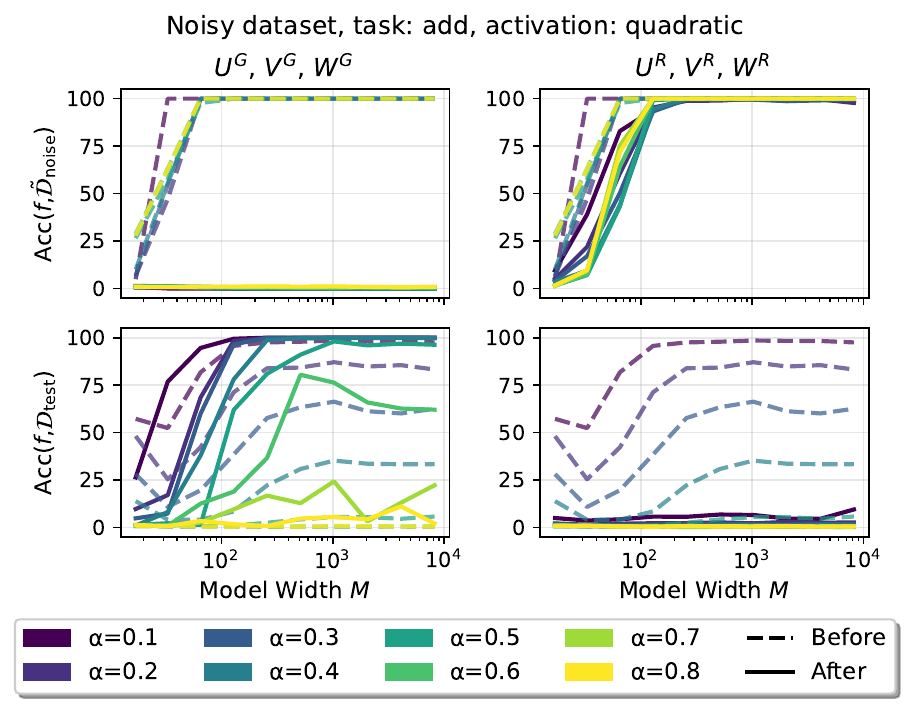}
  \caption{Performance before (dashed) and after (solid) the frequency filtration on models with \textit{quadratic} activation. We remark that frequency filtration also breaks the ceilings of test accuracy.}
  \label{fig:freq_filtration_square}
\end{figure*}

\begin{figure*}[!htbp] 
  \centering
  \includegraphics[width=0.6\linewidth]{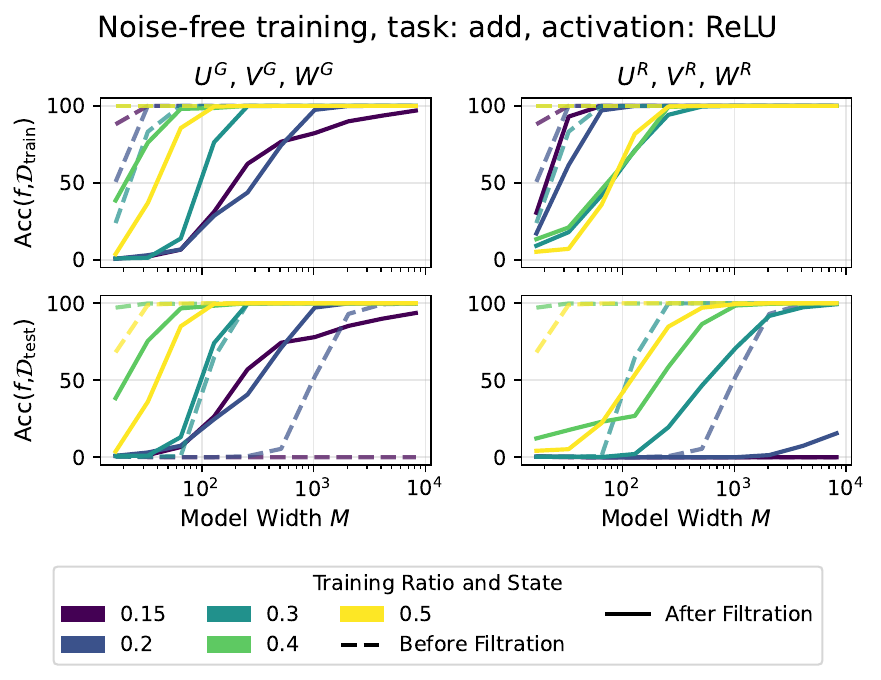}
  \caption{Performance before (dashed) and after (solid) frequency filtration for models trained on the clean modular addition data. 
The models use ReLU activation and are trained with AdamW. 
Overall, test accuracy after filtration ($U^G$, $V^G$, $W^G$) is generally higher, particularly for large models.}
  \label{fig:freq_filtration_clean}
\end{figure*}

Frequency filtration also disentangles generalization and memorization for the modular \textit{subtraction} task (Figure~\ref{fig:freq_filtration_sub}) and for models with \textit{quadratic} activations (Figure~\ref{fig:freq_filtration_square}). 
We observe substantial improvement in generalization after frequency filtration: ReLU models recover test accuracy under severe label noise, while quadratic models exceed previously observed accuracy limits. 
Test accuracy also increases for models trained on noise-free data (Figure~\ref{fig:freq_filtration_clean}).

\paragraph{Frequency filtration on transformers.} To test whether the frequency-based filtering idea is restricted to two-layer networks, we also apply FF to a Transformer trained on modular addition with label noise. The mechanism of the Transformer for the modular addition task on \textit{clean} data has been explored in \citet{nandaprogress}. Specifically, after training, we keep only the top 20 frequencies in the unembedding head and remove the remaining frequency components. This simple intervention substantially reduces the model's ability to memorize noisy labels while improving its test accuracy, suggesting that the frequency-structured rule representation is not exclusive to the two-layer setting. (Table \ref{tab:transformer_frequency_filtration}) These results provide preliminary evidence that FF can extend beyond two-layer networks.

\begin{table}[h]
    \centering
    \caption{Applying frequency filtration to a Transformer trained on modular addition. FF keeps only the top 20 frequencies in the unembedding head.}
    \label{tab:transformer_frequency_filtration}
    \begin{tabular}{lccc}
        \toprule
        Method & Clean Acc. & Noisy Acc. & Test Acc. \\
        \midrule
        Before FF & $100\%$ & $100\%$ & $76.8\%$ \\
        After FF & $98.1\%$ & $36.8\%$ & $87.8\%$ \\
        \bottomrule
    \end{tabular}
\end{table}

\subsection{Relationship among the phases $\varphi^{(a)}_m$, $\varphi^{(b)}_m$, and $\varphi^{(c)}_m$}\label{sec:phase_additional}

Suppose the dominant frequency of neuron $m$ has the form
\begin{subequations}
\begin{align}
& u^G_{mi} = \lambda_u \cos \left( \frac{2\pi}{P}\omega_m i + \varphi^{(a)}_{m,G} \right), \\
& v^G_{mj} = \lambda_v \cos \left( \frac{2\pi}{P}\omega_m j + \varphi^{(b)}_{m,G} \right), \\
& w^G_{mk} = \lambda_w \cos \left( \frac{2\pi}{P}\omega_m k + \varphi^{(c)}_{m,G} \right).
\end{align}
\end{subequations}

For models trained with either ReLU or quadratic activations, there exists $q\in\mathbb{Z}$ such that the following equations are approximately satisfied:
\begin{align}
\varphi^{(a)}_{m,G} + \varphi^{(b)}_{m,G} &= \varphi^{(c)}_{m,G} + 2q\pi && \text{(modular addition)}, \\
\varphi^{(a)}_{m,G} - \varphi^{(b)}_{m,G} &= \varphi^{(c)}_{m,G} + 2q\pi && \text{(modular subtraction)}.
\end{align}

We quantify the empirical deviation of these relationships using the mean squared error (MSE). Specifically, for modular addition:
\begin{equation*}
    \text{Phase MSE} = \frac{1}{M}\sum_{m=1}^M \min_{q\in\mathbb{Z}}\left(\varphi^{(a)}_{m,G} + \varphi^{(b)}_{m,G}-\varphi^{(c)}_{m,G} + 2q\pi\right)^2,
\end{equation*}
and for modular subtraction, the $+$ is replaced by $-$ in the definition above.

\begin{figure*}[!htbp] 
  \centering
  \includegraphics[width=0.7\linewidth]{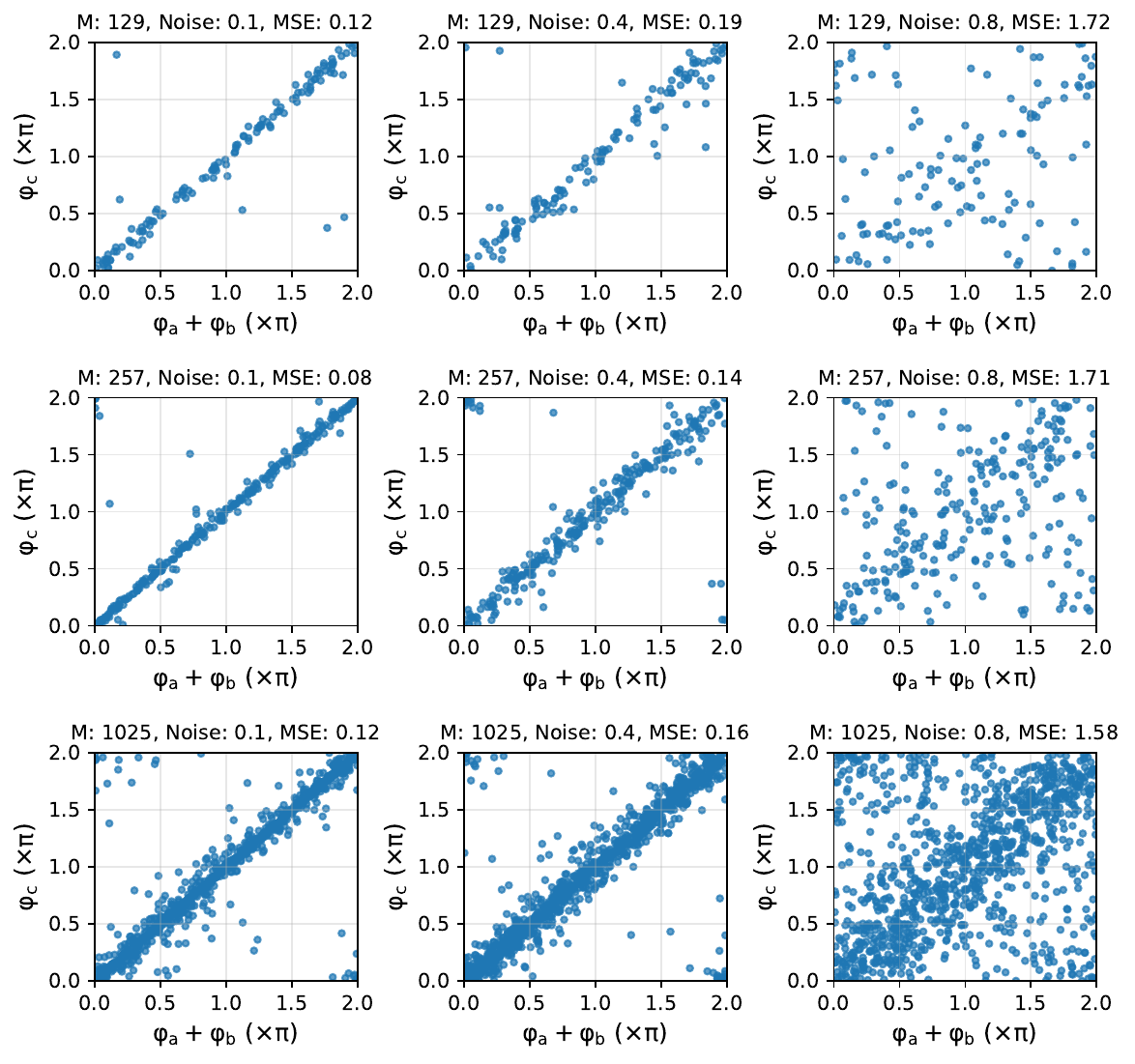}
  \caption{Scatter plot showing the relationship of phases on modular addition tasks. Each point represents a neuron, where $\varphi_a$, $\varphi_b$, and $\varphi_c$ are the phases of $\bm{u}^G$, $\bm{v}^G$, and $\bm{w}^G$, respectively.}
  \label{fig:phases_add_scatter}
\end{figure*}

\begin{figure*}[!htbp] 
  \centering
  \includegraphics[width=0.7\linewidth]{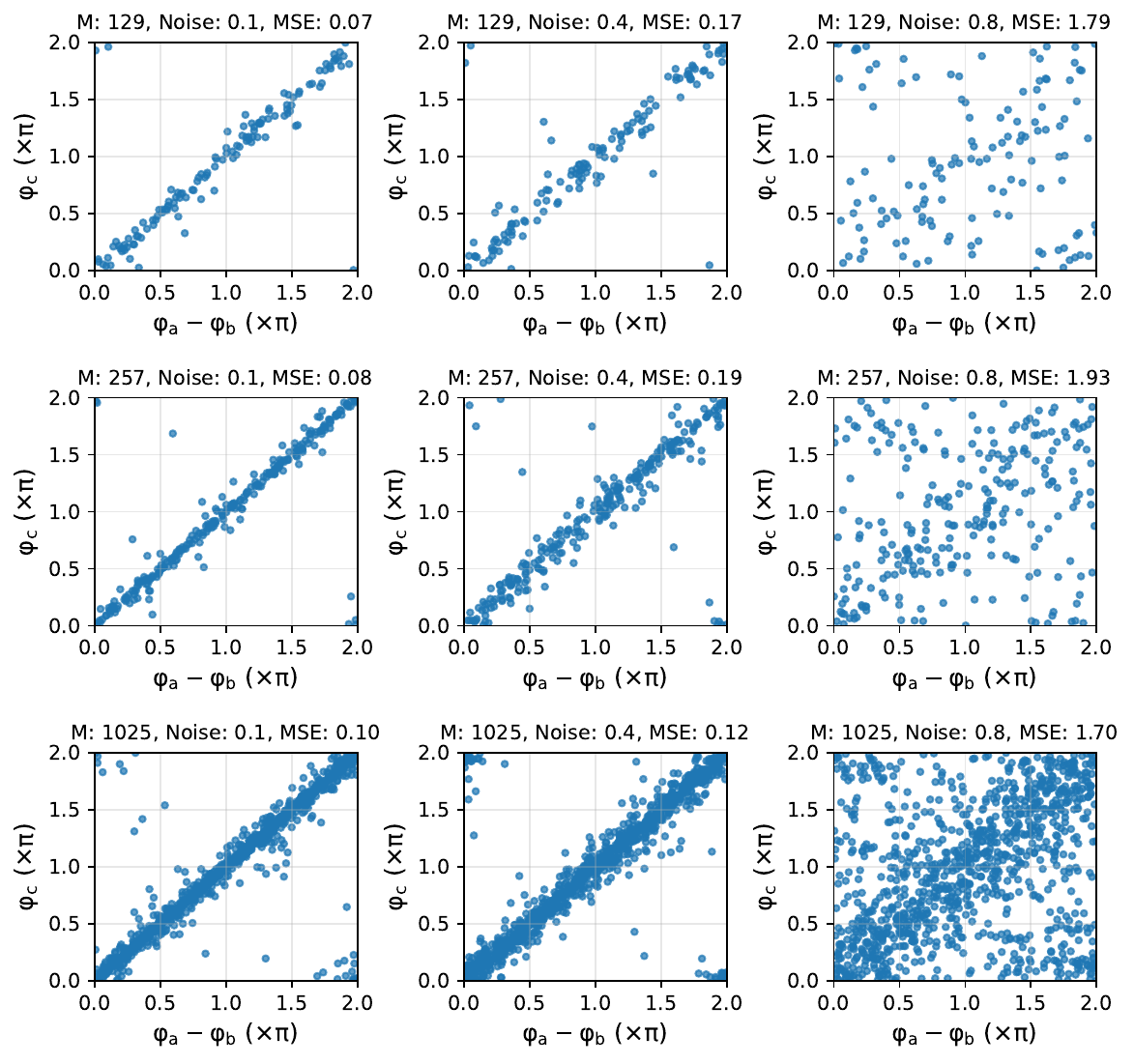}
  \caption{Scatter plot showing the relationship of phases on modular subtraction tasks. Each point represents a neuron, where $\varphi_a$, $\varphi_b$, and $\varphi_c$ are the phases of $\bm{u}^G$, $\bm{v}^G$, and $\bm{w}^G$, respectively.}
  \label{fig:phases_sub_scatter}
\end{figure*}

\begin{figure}[htbp]
    \centering
    \begin{subfigure}{0.45\textwidth}
        \centering
        \includegraphics[width=\textwidth]{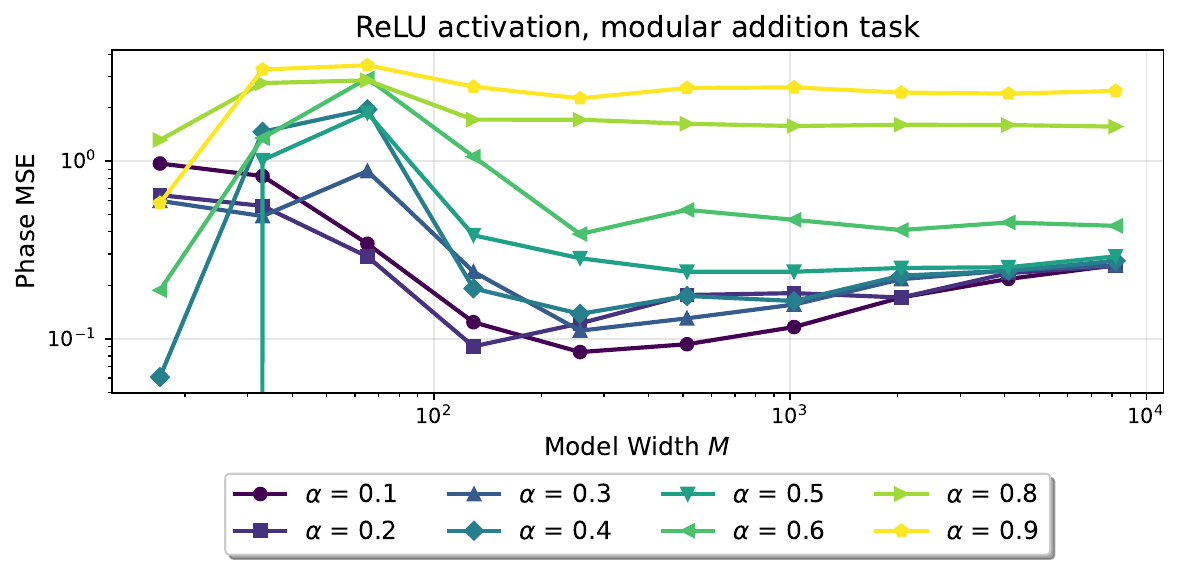}
        \label{fig:sub1}
    \end{subfigure}
    \hfill
    \begin{subfigure}{0.45\textwidth}
        \centering
        \includegraphics[width=\textwidth]{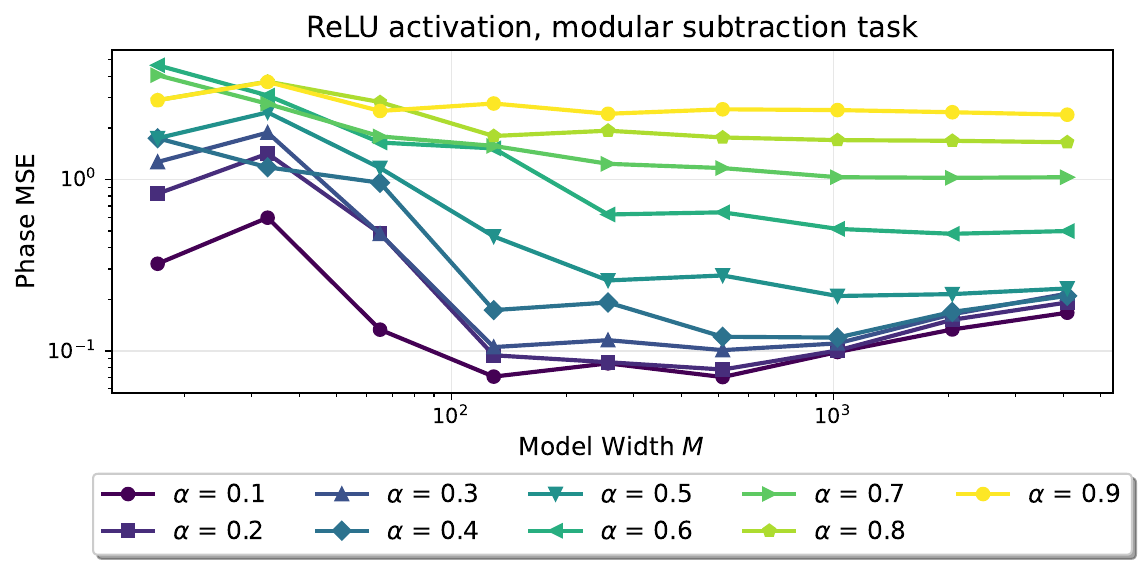}
        \label{fig:sub2}
    \end{subfigure}
    \begin{subfigure}{0.45\textwidth}
        \centering
        \includegraphics[width=\textwidth]{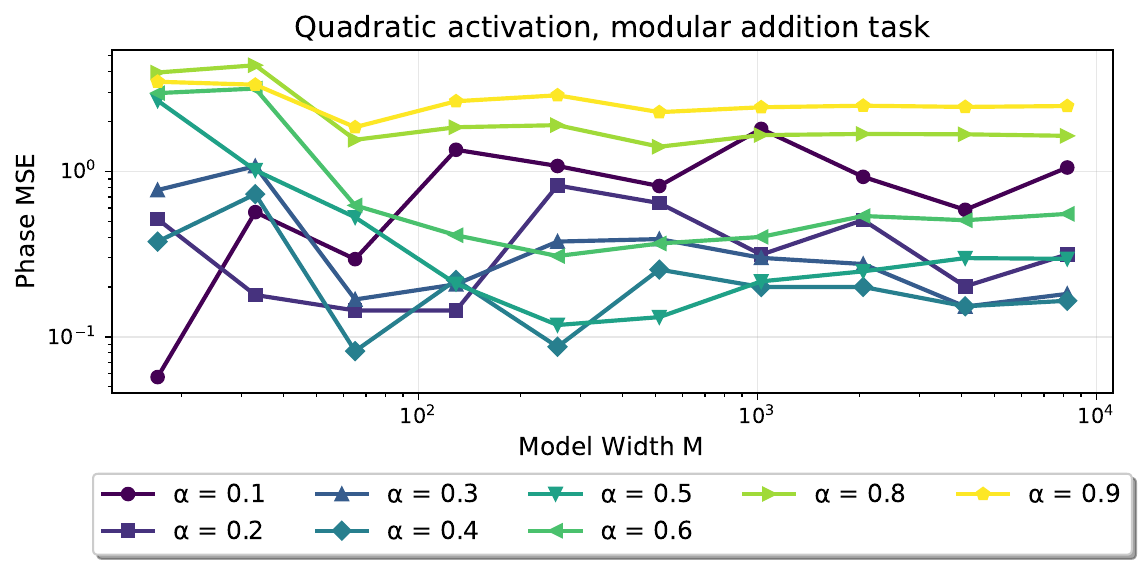}
        \label{fig:sub3}
    \end{subfigure}
    \hfill
    \begin{subfigure}{0.45\textwidth}
        \centering
        \includegraphics[width=\textwidth]{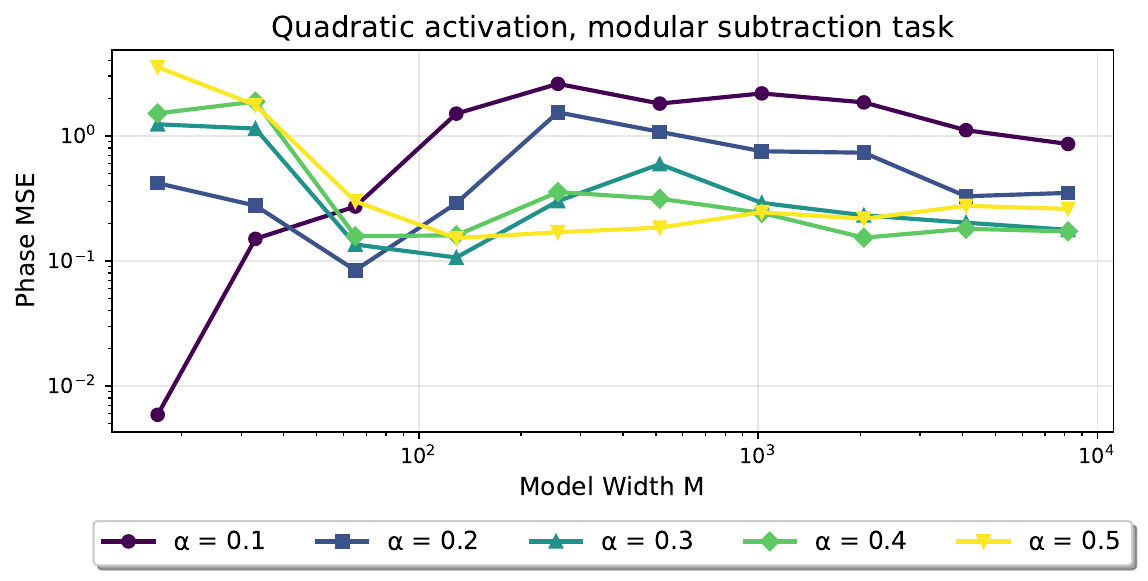}
        \label{fig:sub4}
    \end{subfigure}
    
    \caption{Phase MSE across different tasks, activation functions, model widths, and noise ratios. 
Higher noise ratios increase Phase MSE for ReLU models, and a similar trend is observed for quadratic models on the modular addition task.}
    \label{fig:phase_mse}
\end{figure}

We visualize these relationships in Figures~\ref{fig:phases_add_scatter} (addition) and \ref{fig:phases_sub_scatter} (subtraction), 
and the Phase MSE trends with respect to model size and noise ratio are shown in Figure~\ref{fig:phase_mse}. 
We observe that higher noise ratios generally lead to larger Phase MSE, i.e., larger deviation to the analytical relationship.

\subsection{Uniform distribution of the frequency}\label{sec:uniform_distribution_additional}
\begin{figure*}[!htbp] 
  \centering
  \includegraphics[width=0.75\linewidth]{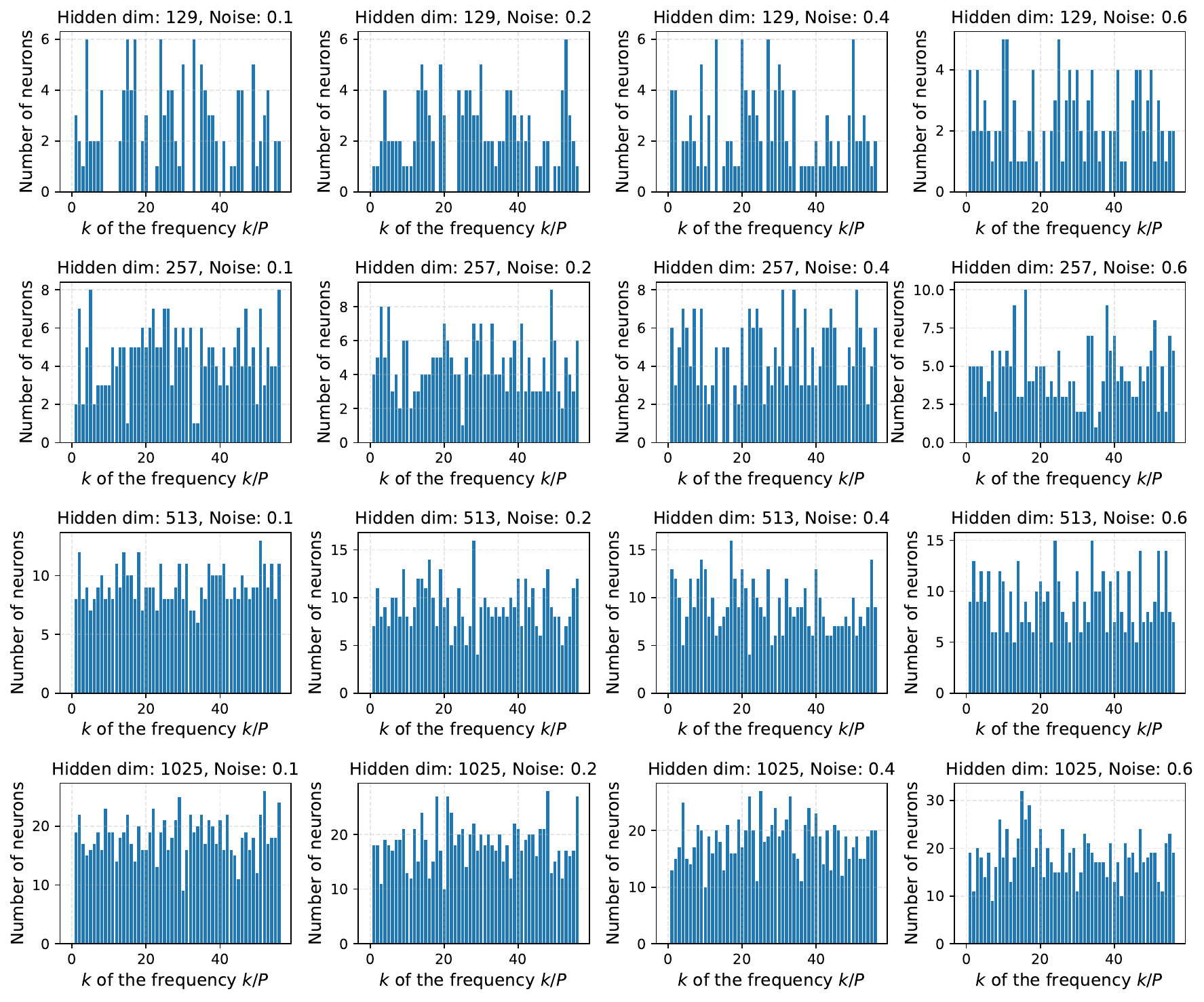}
  \caption{Each frequency $\omega\in\left\{1,\cdots,\frac{P-1}{2}\right\}$ is covered by at least a neuron when the model width is no less than $513$.}
  \label{fig:frequency_distribution}
\end{figure*}

\begin{figure*}[!htbp] 
  \centering
  \includegraphics[width=0.95\linewidth]{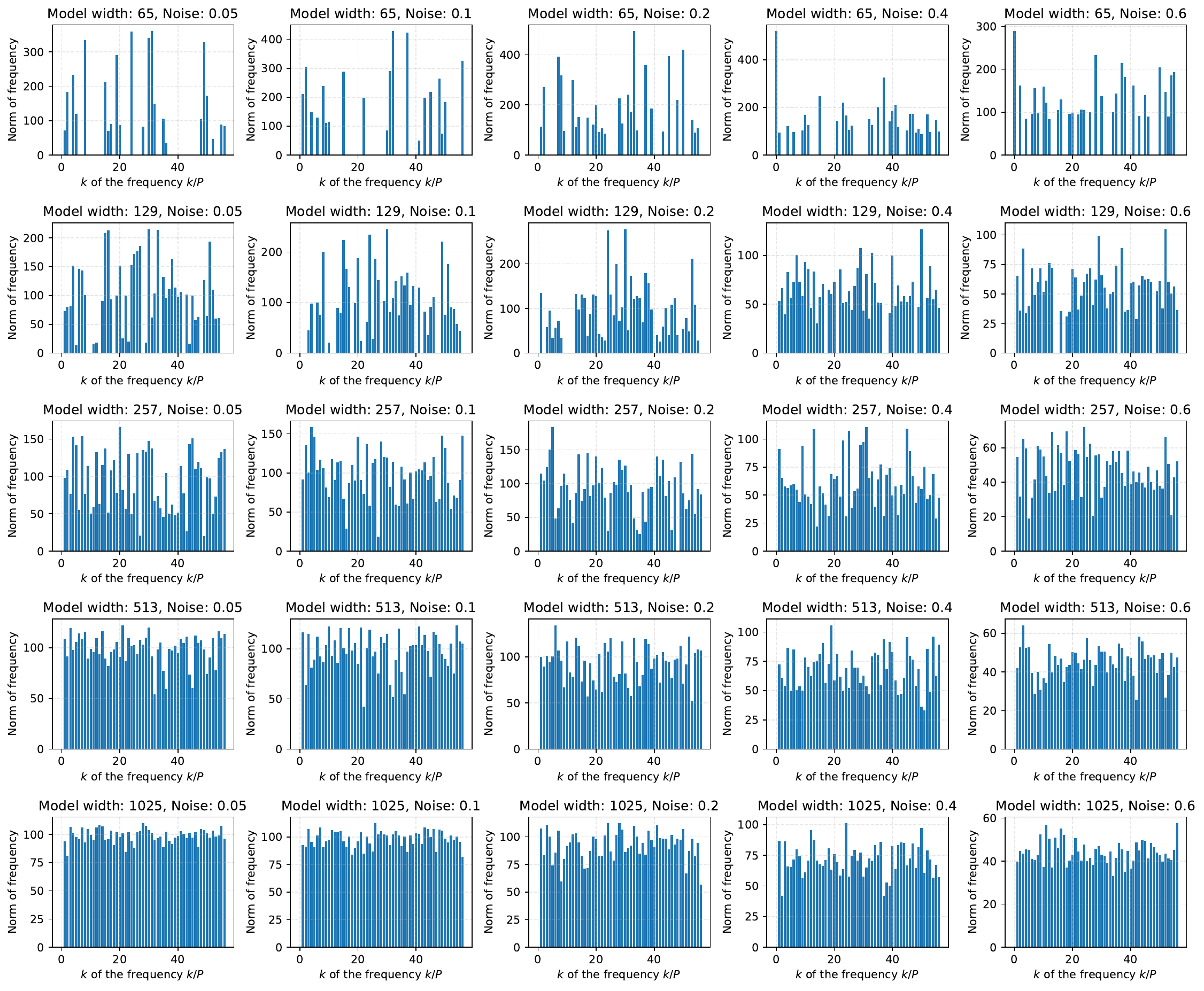}
  \caption{The magnitude (norm) of each frequency $\omega\in\left\{1,\cdots,\frac{P-1}{2}\right\}$ of $\bm{W}^G$ after frequency filtration to each neuron.}
  \label{fig:frequency_norm}
\end{figure*}

Figure \ref{fig:frequency_distribution} shows that each frequency is covered by at least one neuron after frequency filtration (only dominant frequency left for each neuron).  Figure \ref{fig:frequency_norm} also shows that the distribution of the frequency magnitude after filtration is close to uniform at larger model width. Here, let $\tilde{\bm{W}}^G$ is the Fourier transform of $\bm{W}^G$ with each column constituted by $\left\{\bm{w}_m\right\}_{m=1}^M$, so Figure \ref{fig:frequency_norm} plots the norm of each row vector in $\tilde{\bm{W}}^G$ with index in $\left\{1,\cdots,\frac{P-1}{2}+1\right\}$. 

\section{Supplementary Materials for Section \ref{sec:network-decomposition}}\label{sec:additional-network-decomposition}

\subsection{Performance of Algorithm \ref{alg:neuron_selection} using Str. or IPR}

\begin{figure*}[!htbp] 
  \centering
  \includegraphics[width=0.8\linewidth]{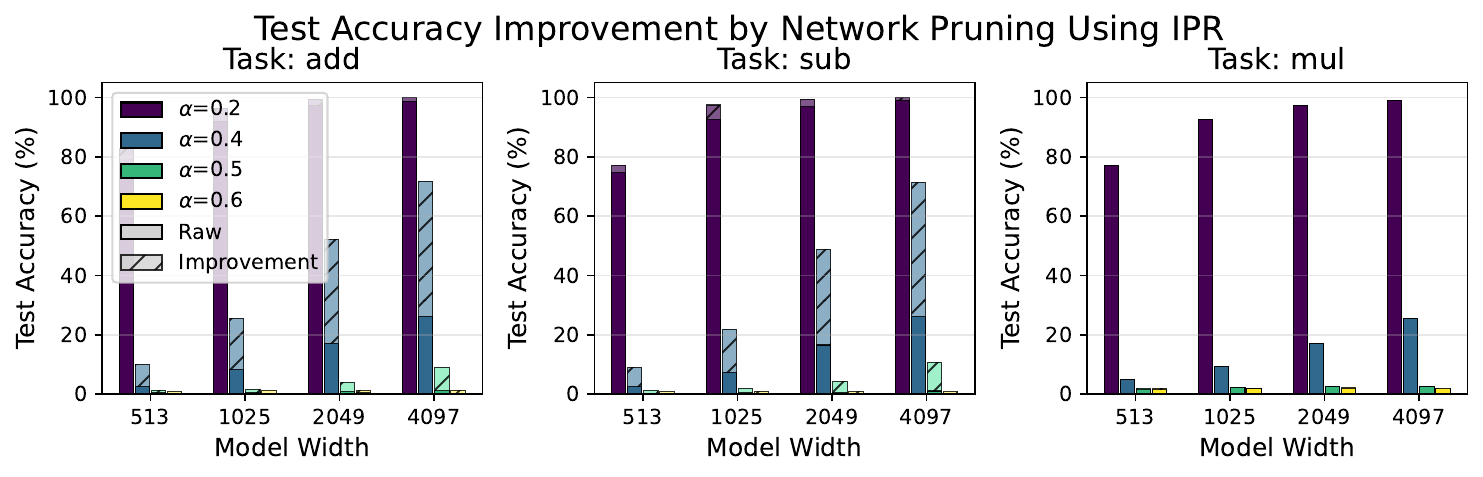}
  \caption{Generalization improvement across various tasks by neuron selection/pruning using IPR. 
  No noticeable improvement is observed for the modular multiplication task.}
  \label{fig:neuron_selection_improvement_IPR}
\end{figure*}

\begin{figure*}[!htbp] 
  \centering
  \includegraphics[width=0.8\linewidth]{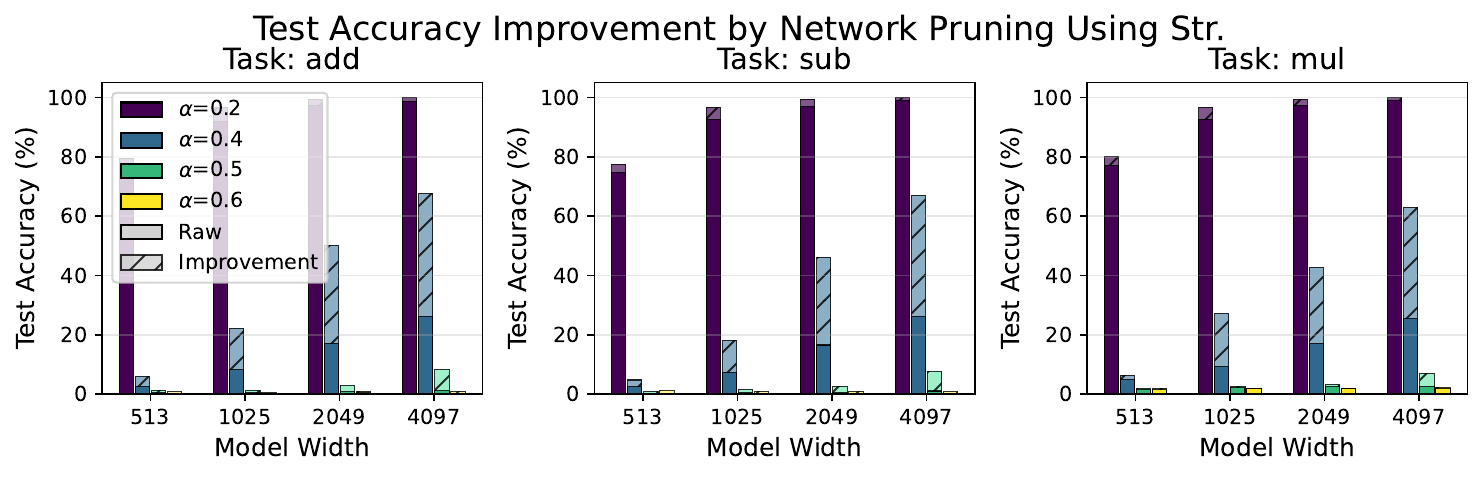}
  \caption{Generalization improvement across various tasks by neuron selection/pruning using Str. 
  Significant improvement is also observed for the modular multiplication task.}
  \label{fig:neuron_selection_improvement_Str}
\end{figure*}

In this section, we examine generalization improvements across different tasks achieved through neuron selection. 
Figure~\ref{fig:neuron_selection_improvement_IPR} reports the improvement obtained using IPR for neuron selection, 
while Figure~\ref{fig:neuron_selection_improvement_Str} presents the corresponding results for Str.

Comparing the results on the modular multiplication task illustrates the generality of the method using Str. 
This task lacks periodic patterns relevant for generalization, so measuring neuron importance via IPR offers limited improvement.


\newpage

\end{document}